\documentclass[preprint,12pt]{elsarticle}

\usepackage[utf8]{inputenc}
\usepackage[english]{babel}
\usepackage{amsmath,amssymb}
\usepackage{graphicx}
\usepackage[hidelinks]{hyperref}
\usepackage{booktabs}
\usepackage{multirow}
\usepackage{array}
\usepackage{caption}
\usepackage{subcaption}
\usepackage{setspace}
\usepackage{xcolor}
\usepackage{rotating}
\usepackage{tabularx}
\usepackage{makecell}
\usepackage{xurl}
\usepackage{placeins}
\usepackage[switch]{lineno}

\journal{Computers and Electronics in Agriculture}

\begin{document}

\begin{frontmatter}

\title{AgrI Challenge: A Data-Centric AI Competition for Cross-Team Validation in Agricultural Vision}

\author[aff1]{Mohammed Brahimi\corref{cor1}}
\author[aff2]{Karim Laabassi}
\author[aff1]{Mohamed Seghir Hadj Ameur\fnref{eq1}}
\author[aff1]{Aicha Boutorh\fnref{eq1}}
\author[aff2]{Badia Siab-Farsi\fnref{eq1}}
\author[aff3]{Amin Khouani\fnref{eq1}}
\author[aff1]{Omar Farouk Zouak}
\author[aff1]{Seif Eddine Bouziane}
\author[aff4]{Kheira Lakhdari}
\author[aff2]{Abdelkader Nabil Benghanem}

\cortext[cor1]{Corresponding author}

\fntext[eq1]{These authors contributed equally to this work.}

\fntext[em]{Emails: mohamed.brahimi@ensia.edu.dz, 
karim.laabassi@edu.ensa.dz,
mohamed.hadj.ameur@ensia.edu.dz,
aicha.boutorh@ensia.edu.dz,  
badia.farsi@edu.ensa.dz,
a\_khouani@esi.dz, 
omar.zouak@ensia.edu.dz,
seifeddine.bouziane@ensia.edu.dz, 
keira.lakhdari@ensta.edu.dz, 
nabil.benghanem@edu.ensa.dz}

\address[aff1]{The National School of Artificial Intelligence (ENSIA), Algiers, 16046, Algeria}
\address[aff2]{National Higher School of Agronomy (ENSA), Algiers, 16051, Algeria}
\address[aff3]{National Higher School of Computer Science (ESI), Algiers, 16000, Algeria}
\address[aff4]{National Higher School of Advanced Technologies (ENSTA), Algiers, 16000, Algeria}

\begin{abstract}

Machine learning models in agricultural vision often achieve high accuracy on curated datasets but fail to generalize under real field conditions due to distribution shifts between training and deployment environments. Moreover, most machine learning competitions focus primarily on model design while treating datasets as fixed resources, leaving the role of data collection practices in model generalization largely unexplored.

We introduce the \textit{AgrI Challenge}, a data-centric competition framework in which multiple teams independently collect field datasets, producing a heterogeneous multi-source benchmark that reflects realistic variability in acquisition conditions. To systematically evaluate cross-domain generalization across independently collected datasets, we propose \textit{Cross-Team Validation} (CTV), an evaluation paradigm that treats each team’s dataset as a distinct domain. CTV includes two complementary protocols: \textit{Train-on-One-Team-Only} (TOTO), which measures single-source generalization, and \textit{Leave-One-Team-Out} (LOTO), which evaluates collaborative multi-source training.

Experiments reveal substantial generalization gaps under single-source training: models achieve near-perfect validation accuracy yet exhibit validation–test gaps of up to 16.20\% (DenseNet121) and 11.37\% (Swin Transformer) when evaluated on datasets collected by other teams. In contrast, collaborative multi-source training dramatically improves robustness, reducing the gap to 2.82\% and 1.78\%, respectively.

The challenge also produced a publicly available dataset of 50,673 field images of six tree species collected by twelve independent teams, providing a diverse benchmark for studying domain shift and data-centric learning in agricultural vision.

\end{abstract}


\end{frontmatter}

\newpage

\section{Introduction}

Machine learning has achieved remarkable advances in agricultural applications, particularly in plant recognition, enabling systems to automatically identify plant species, monitor crop health, and support tasks ranging from crop disease detection to yield prediction. However, a persistent challenge undermines practical deployment: models that perform exceptionally well on benchmark datasets often fail dramatically when applied to real-world field conditions. This generalization gap is exemplified by models trained on the widely-used PlantVillage dataset, which achieve over 99\% accuracy in controlled settings but drop to 54\% when deployed in actual farm environments \citep{hughes2015open, barbedo2018impact,Benabbas2024,Benabbas2025}.

Traditional machine-learning competitions have helped advance algorithms, but they also reinforce a key limitation: participants are given fixed datasets prepared by organizers. As a result, competitors focus on improving model design, tuning parameters, and training methods, while the data itself stays unchanged. This encourages a model-centered view in which datasets are treated as static resources instead of parts of the learning process that can be improved over time. Important aspects such as how data is collected, sampled, and made diverse, factors that strongly affect real-world performance, are therefore rarely studied \citep{competition2024,data-centric2024}.

Recent work in data-centric AI has begun to address this limitation by shifting focus from model optimization to data quality improvement \citep{mazumder2023dataperf}. Initiatives such as the Data-Centric AI Competition organized by DeepLearning.AI demonstrated that systematic dataset refinement can yield performance gains exceeding 20\%, even with fixed model architectures. However, these efforts still operate within a closed dataset framework, improving existing data rather than exploring how different collection methodologies create distributional shifts that challenge model generalization.

This study introduces the AgrI Challenge, a multi-phase competition framework that integrates participant-led field data collection with collaborative model development. Unlike conventional competitions where data is provided, AgrI Challenge participants collect their own agricultural data over a dedicated two-day field period, then develop models trained exclusively on their collected datasets. This design forces engagement with the complete machine learning pipeline and generates datasets that reflect diverse environmental conditions, device characteristics, and sampling strategies.

Beyond its research objectives, the AgrI Challenge also serves a pedagogical role by exposing students to the full end-to-end machine learning workflow, from data collection and curation to model training and evaluation. This contrasts with traditional coursework that often emphasizes model development on pre-existing datasets, enabling participants to gain practical experience with the complete lifecycle of real-world AI systems.

To evaluate this framework, we develop Cross-Team Validation (CTV), a novel evaluation paradigm where datasets collected by different teams serve as distinct domains for systematic generalization assessment. CTV treats each team's dataset as representing a unique combination of collection methodology, environmental context, and device characteristics, capturing authentic inter-domain variation that simulates real-world deployment scenarios.

We implement two specific CTV protocols: Leave-One-Team-Out (LOTO), where models are trained on aggregated data from $n-1$ teams and tested on the held-out team's data, evaluating collaborative training scenarios; and Train-on-One-Team-Only (TOTO), where models are trained on individual team datasets and tested on independent data, representing competitive scenarios with data siloing.

This study addresses three research questions:

\begin{enumerate}
    \item How well do agricultural vision models generalize across independently collected field datasets?
    \item Does collaborative multi-source training substantially improve robustness compared to single-source training?
    \item Can cross-team evaluation reveal dataset quality differences?
\end{enumerate}

We provide dual architecture baselines, DenseNet121 (CNN) and Swin Transformer, enabling future dataset users to benchmark against established backbones.

The remainder of this paper is organized as follows: Section 2 reviews related work; Section 3 describes the AgrI Challenge framework, dataset characteristics, and CTV methodology; Section 4 presents results; Section 5 discusses findings; and Section 6 concludes.

\FloatBarrier

\section{Related Work}
This section reviews prior research relevant to the present study. We first discuss the evolution from model-centric to data-centric AI and its implications for machine learning research. We then examine existing agricultural AI datasets and competitions that have shaped benchmarking practices in this domain. Finally, we position the AgrI Challenge within this landscape, highlighting how it extends current approaches through participant-led data collection and cross-team evaluation.

\subsection{From Model-Centric to Data-Centric AI}

Machine learning research has traditionally centered on developing and optimizing model architectures using fixed, publicly available datasets. Large-scale competitions, such as those hosted on Kaggle, have been instrumental in advancing model performance by crowdsourcing innovative modeling strategies under standardized conditions \citep{kaggle2023competitions}. However, this model-centric paradigm treats datasets as static, immutable resources, often sidelining crucial aspects of the machine learning pipeline such as data collection, cleaning, and augmentation \citep{murindanyi2024enhanced}.

The limitations of this approach are increasingly evident in high-stakes applications. \citep{sambasivan2021everyone} documented the phenomenon of ``data cascades,'' where 92\% of practitioners experience systematic flaws in data quality that compound through the development pipeline. In the agricultural domain, these flaws often manifest as "shortcut learning"; for instance, \citep{barbedo2018impact} demonstrated that models trained on laboratory-collected datasets frequently learn to identify background patterns or lighting conditions rather than the relevant plant characteristics.

To address these shortcomings, the data-centric AI movement proposes a paradigm shift where data quality, diversity, and representativeness become the primary targets of optimization \citep{ng2021datacentric}. This transition is being codified through new benchmarking standards. \citep{mazumder2023dataperf} introduced DataPerf, a suite designed to facilitate data-centric innovation, while the 2021 Data-Centric AI Competition demonstrated that systematic dataset improvement could yield performance gains exceeding 20\% while keeping model architectures fixed.

Despite these advancements, many early data-centric competitions operated within closed frameworks that did not fully account for generalization across independent, real-world data sources. Recent research has therefore emphasized the creation of realistic, field-collected datasets, such as "Tomato-Village" \citep{gehlot2023tomato} and "CropDP-181" \citep{kong2022spatial}, which capture the inherent complexity of natural environments, including background noise, varying illumination, and co-occurring diseases. This evolution further aligns with the principles of "Participatory Machine Learning," which integrates crowdsourcing and field-based collection to leverage human expertise.

Ultimately, the agricultural sector, characterized by extreme variability in environmental conditions and crop phenotypes, represents a critical frontier for data-centric AI. Challenges in plant classification, pest detection, and yield prediction are increasingly defined not by algorithmic innovation alone, but by the quality and representativeness of the underlying data \citep{yu2023progress, lin2023effective, kong2022spatial}. This shift acknowledges that the robustness of AI in the field is often limited by the data’s ability to reflect infield complexities rather than the complexity of the model architecture itself \citep{gehlot2023tomato}.

\subsection{Agricultural AI Competitions and Datasets}

The trajectory of agricultural machine learning has been significantly shaped by public competitions and the release of large-scale datasets. Early benchmarks focused primarily on classification tasks under controlled conditions. For instance, the AI Challenger 2018 competition catalyzed research into crop disease classification using a dataset of over 50,000 images across 61 classes \citep{lin2023effective}. Similarly, the Cassava Leaf Disease Classification challenge provided a foundation for identifying health states in staple crops \citep{maryum2021cassava}, while the PlantCLEF series has consistently pushed the boundaries of large-scale plant identification \citep{goeau2022overview, goeau2023overview}. More recently, remote sensing has gained prominence, exemplified by the "Tianchi Competition" and its associated Barley Remote Sensing Dataset (BRSD), which leverages UAV imagery for crop-type classification \citep{wu2025edge}. Beyond disease and species identification, varietal-level classification has also benefited from curated datasets, with studies leveraging over 31,000 scanner-captured wheat grain images and transfer learning to achieve accuracies exceeding 95\% \citep{laabassi2021wheat}.

Despite the success of these benchmarks, a critical limitation persists regarding laboratory-to-field generalization. The PlantVillage dataset \citep{hughes2015open}, while foundational, serves as a primary example of this gap. Although models trained on its 54,000+ images often achieve near-perfect accuracy on internal test sets, their performance frequently collapses when deployed in natural environments. Studies have shown that accuracy can decrease significantly from over 90\% to below 60\% due to the absence of ``in-the-wild'' variables such as complex backgrounds and varying illumination \citep{barbedo2018impact, gehlot2023tomato}.

To bridge this gap, a new generation of datasets has emerged to capture "in-field" complexities: Tomato-Village, specifically designed for disease detection within natural, unconstrained environments \citep{gehlot2023tomato}, and CropDP-181 that integrates data directly from in-field IoT monitoring systems to reflect real-world phenological variations \citep{kong2022spatial}.

A significant transformation is currently underway in agricultural AI hackathons, shifting the focus from architectural modification to data-centric strategies. These modern competitions emphasize data curation, quality control, and multimodal fusion (Table \ref{tab:compagri}). Unlike traditional challenges, these events require participants to optimize the data processing pipeline, addressing noise, class imbalance, and sensor calibration, to ensure robust performance in the field.

\begin{table}[!htbp]
\centering
\footnotesize
\setlength{\tabcolsep}{4pt}
\caption{Summary of Data Strategies and Impact in Agricultural Machine Learning}
\label{tab:compagri}
\begin{tabularx}{\textwidth}{@{} l >{\raggedright\arraybackslash}X >{\raggedright\arraybackslash}X >{\raggedright\arraybackslash}p{3.5cm} @{}}
\toprule
\textbf{Competition/Paper} & \textbf{Data Strategies} & \textbf{Evaluation Impact} & \textbf{Reference} \\
\midrule
Agriculture-Vision & Label correction and augmentation & +8\% F1 score improvement on anomalies & \href{https://www.agriculture-vision.com}{agriculture-vision} \citep{chiu2020agriculture} \\
\addlinespace
AI for Good & Sensor fusion and outlier removal & RMSE reduction on yield forecasting & \citep{esteva2023datacentric} \\
\addlinespace
Tropical Agri Mapping & Active learning and SSL pretraining & Scalable performance in cloud-covered regions & \citep{pinto2025data} \\
\addlinespace
Farm Robotics (2025) & Cleaning heterogeneous in-field datasets & Robust generalization across diverse farms & \url{https://www.farmroboticschallenge.ai/2025results} \\
\addlinespace
Beyond Visible Spectrum & Preprocessing chains for spectral noise & Improved accuracy via data augmentation & \url{https://www.kaggle.com/competitions/beyond-visible-spectrum-ai-for-agriculture-2025p2} \\
\bottomrule
\end{tabularx}
\end{table}

This evolution acknowledges that in the high-variability domain of agriculture, the "intelligence" of a system is increasingly derived from the representativeness and cleanliness of the training data rather than the depth of the neural network architecture alone.


\subsection{Positioning of AgrI Challenge}

The AgrI Challenge extends the data-centric paradigm by integrating participant-led data collection with collaborative model development. Table~\ref{tab:framework_comparison} positions AgrI Challenge relative to prior work.

Existing frameworks largely follow one of two patterns. Centralized benchmarks such as PlantVillage \citep{hughes2015open} and AI Challenger 2018 \citep{lin2023effective} focus on model accuracy within a competitive setting, producing static datasets that participants consume but do not shape. DataPerf \citep{mazumder2023dataperf} shifts attention toward dataset quality, yet operates on pre-provided, single-domain data. CRDDC-2022 \citep{arya2022crowdsensing} improves geographic diversity through multi-country collection, but remains centralized and purely competitive.

AgrI Challenge distinguishes itself by combining participant-led field collection, cross-team evaluation (CTV), and a collaborative-competitive structure. This design ensures that data diversity, generalization across teams, and dataset quality are treated as first-class objectives of the competition process.

\begin{table}[!htbp]
\centering
\caption{Positioning of AgrI Challenge relative to prior agricultural AI frameworks}
\label{tab:framework_comparison}
\small
\begin{tabularx}{\columnwidth}{lXX}
\toprule
\textbf{Framework} & \textbf{Data Source} & \textbf{Evaluation Scope} \\
\midrule
PlantVillage
& Organizer (lab)
& Single-domain \\

AI Challenger 2018
& Organizer (field)
& Single-domain \\

DataPerf
& Pre-provided dataset
& Single-domain \\

CRDDC-2022
& Organizer (multi-country)
& Cross-domain \\

\textbf{AgrI Challenge}
& \textbf{Participant-led field collection}
& \textbf{Cross-domain (CTV); emergent multi-source academic dataset} \\
\bottomrule
\end{tabularx}
\end{table}

\section{Materials and Methods}
This section describes the methodological framework used in this study, including dataset construction, preprocessing procedures, evaluation protocols, and experimental setup. It details how the AgrI Challenge dataset was collected and curated, the strategies used to ensure data quality and cross-team independence, and the experimental design adopted to evaluate model performance and generalization. Together, these components establish a reproducible pipeline for assessing the proposed data-centric framework.
 
\subsection{AgrI Challenge Framework Design}
 
The \textbf{AgrI Challenge} was designed as an interdisciplinary framework to address the complexities of tree species classification through a data-centric machine learning pipeline. The initiative involved \textbf{12 collection groups} (\textbf{11 participant teams plus the organizing committee}). Each participant team comprised five students from different regions of Algeria. The task focused on \textbf{tree species classification} across six classes.
 
To ensure both ecological validity of annotations and technical robustness of models, each participating team was required to be \textbf{interdisciplinary}, combining students with computing backgrounds (computer science, artificial intelligence, data science) and students with ecological or agronomic backgrounds (agriculture, forestry, plant sciences). In addition to the participating teams, the organizing team responsible for coordinating the AgrI Challenge also contributed to the data collection process. This mixed composition ensured high-quality species labelling grounded in domain expertise while supporting effective machine learning system design and implementation.
 
\begin{figure}[!htbp]
    \centering
    \includegraphics[width=\textwidth]{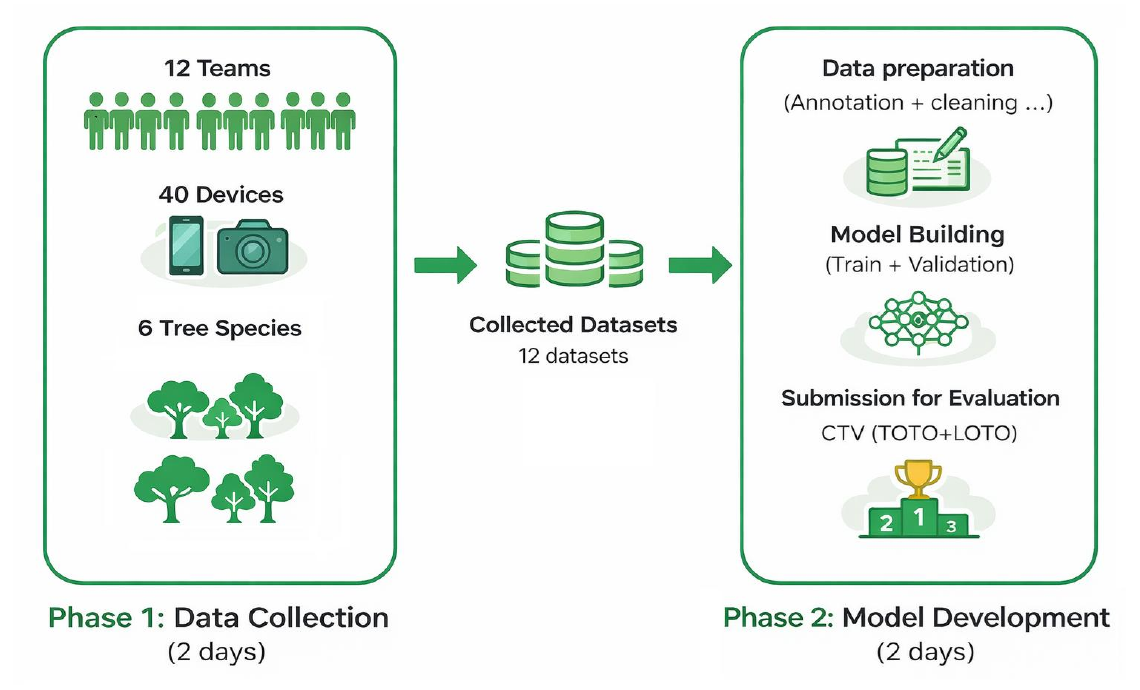}
    \caption{AgrI Challenge workflow showing the two-phase competition design and Cross-Team Validation evaluation protocols.}
    \label{fig:workflow}
\end{figure}

The competition design followed a two-phase protocol (Figure~\ref{fig:workflow}):
\begin{enumerate}

     \item \textbf{Data Collection Phase (2 days):} Teams collected field data with expert mentorship at the experimental and teaching facilities of ENSA,\footnote{\textit{National Higher School of Agronomy, Algiers, Algeria (ENSA)} is a long-established and widely recognised agronomic institution in Algeria (founded in 1905).} ensuring access to representative agro-ecosystems and well-maintained botanical collections. Teams had freedom to choose collection devices and experimental design, including sampling strategies, environmental coverage, and imaging protocols. The plant species included in the study were selected and validated by ENSA domain experts to ensure taxonomic reliability and ecological relevance.
    
    \item \textbf{Model Development Phase (2 days):} Utilizing the computational facilities at ENSIA\footnote{\textit{The National School of Artificial Intelligence (ENSIA), Algiers, Algeria} is a national centre of excellence dedicated to education and research in artificial intelligence and data science.}, teams performed data annotation and preprocessing, then used their collected data to train classification models. Teams were mentored by ENSIA specialists in AI and machine learning, who provided technical guidance on data preparation, model design, training, and evaluation.

\end{enumerate}

Beyond AgrI challenge research objectives, the competition was intentionally designed as a pedagogical exercise, encouraging students to engage with the complete data-centric machine learning pipeline, including data acquisition, dataset preparation, and model development.

\subsection{Plant Material and Target Species}
 
This subsection describes the botanical material used in the AgrI Challenge. We first present the species selection rationale and taxonomic framework, then summarise the visual properties relevant to automated classification.
 
\subsubsection{Species Selection and Taxonomic Framework}
\label{sec:species_selection}
 
Six species from the living collections of the Botanical Garden of the National Higher School of Agronomy (ENSA) were selected to constitute a botanical reference dataset aimed at training and evaluating digital plant recognition models. The selection of these species, which are part of the plant diversity of ENSA, combines \textbf{agronomic relevance}, \textbf{ornamental purpose}, \textbf{ecological interest}, and \textbf{morphological richness} to facilitate the development of datasets suitable for computer vision applications.
 
The resulting dataset encompasses both indigenous Algerian species and two ornamental introductions. The four indigenous species are: Carob tree (\textit{Ceratonia siliqua} L., Caroubier), Narrow-leaved ash (\textit{Fraxinus angustifolia} Vahl, Fr\^ene \`a feuilles \'etroites), Atlas pistachio (\textit{Pistacia atlantica} Desf., Pistachier de l'Atlas), and Oak (\textit{Quercus} spp., Ch\^ene). The two ornamental introductions are Tipu tree (\textit{Tipuana tipu} (Benth.) Kuntze, Tipu), located along a tree-lined avenue, and Peruvian pepper tree (\textit{Schinus molle} L., Faux poivrier), situated near a water feature.
 
Taxonomic nomenclature for native species was standardized according to the African Plant Database\footnote{\url{https://africanplantdatabase.ch}
, last accessed 2026.}, while nomenclature for exotic species followed Plants of the World Online\footnote{\url{https://powo.science.kew.org}
, last accessed 2026.}.
 
\paragraph{Quercus super-class}
\label{sec:quercus_grouping}
A critical architectural decision in the dataset was the definition of the \textit{Quercus} super-class, grouping \textit{Quercus ilex} L.\ (Ch\^ene vert), \textit{Quercus suber} L.\ (Ch\^ene-li\`ege), and \textit{Quercus canariensis} Willd.\ (Ch\^ene z\'een) into a single systematic unit based on:
\begin{itemize}
    \item \textbf{Sample scarcity:} Insufficient per-species specimen counts within the botanical garden to support robust, high-granularity model training.
    \item \textbf{Ecological coherence:} The phylogenetic proximity and co-occurrence of these species across Algerian forest biomes~\citep{quezel2002}.
\end{itemize}

\subsubsection{Visual Properties of Target Classes}
\label{sec:visual_cues}
 
For automated image classification, recognition relies on three temporally stable visual channels: \textbf{leaf morphology}, \textbf{trunk texture}, and \textbf{canopy architecture}. While reproductive structures (flowers and fruits) are standard in classical botany, they were excluded from the classification scope due to their \textbf{fugacious} (seasonally ephemeral) appearance, which would introduce temporal acquisition bias and limit model generalisation outside of specific seasons. The three retained visual channels (leaves, trunk, and canopy) are observable year-round and thus provide a temporally stable recognition signal.

The dataset is characterised by a structural asymmetry in leaf architecture: five classes exhibit \textbf{compound leaves} with pinnate arrangements, while the \textbf{\textit{Quercus} super-class} is distinguished by \textbf{simple leaves}. This requires models to differentiate between subtler features like leaflet count and symmetry for compound-leaf species, versus margin geometry for the simple-leaf group.

Table~\ref{tab:visual_cues} summarises the primary and secondary visual features for each class.
 
\begin{table}[!ht]
\centering
\caption{Visual and morphological properties of the six target species.}
\label{tab:visual_cues}
\small
\renewcommand{\arraystretch}{1.4}
\begin{tabular}{@{}p{3.2cm}p{4.8cm}p{4.8cm}@{}}
\toprule
\textbf{Class / Species} & \textbf{Primary Visual Feature (Leaf Morphology)} & \textbf{Secondary Feature (Trunk \& Canopy)} \\
\midrule
\textit{Ceratonia siliqua} \par {\small (Carob)}
& Compound, paripinnate; 2--5 pairs of thick, leathery leaflets
& Globular canopy; brown fibrous pods
\\[4pt] \hline \\[-6pt]
\textit{Fraxinus angustifolia} \par {\small (Narrow-leaved ash)}
& Compound, opposite pinnate; 5--13 narrow, lanceolate leaflets
& Slender, upright habit; elongated winged samaras
\\[4pt] \hline \\[-6pt]
\textit{Pistacia atlantica} \par {\small (Atlas pistachio)}
& Compound, pinnate; 7--11 leaflets; foliage turns red in autumn
& Smooth grey-green trunk; dense, compact crown
\\[4pt] \hline \\[-6pt]
\textit{Quercus} spp. \par {\small (Oak super-class)}
& Simple leaves; variable margin geometry (entire, lobed, or serrate)
& Thick, deeply fissured trunk; \textit{Q.~suber} exhibits a thick corky layer
\\[4pt] \hline \\[-6pt]
\textit{Schinus molle} \par {\small (Peruvian pepper)}
& Fine, drooping compound pinnate leaflets; pendulous branches
& Weeping canopy silhouette; clusters of small pink berries
\\[4pt] \hline \\[-6pt]
\textit{Tipuana tipu} \par {\small (Tipu tree)}
& Large compound pinnate leaves; numerous oblong leaflets
& Broad, spreading canopy; brown-orange trunk hue
\\
\bottomrule
\end{tabular}
\end{table}

\begin{figure}[!htbp]
    \centering
    \includegraphics[width=\textwidth]{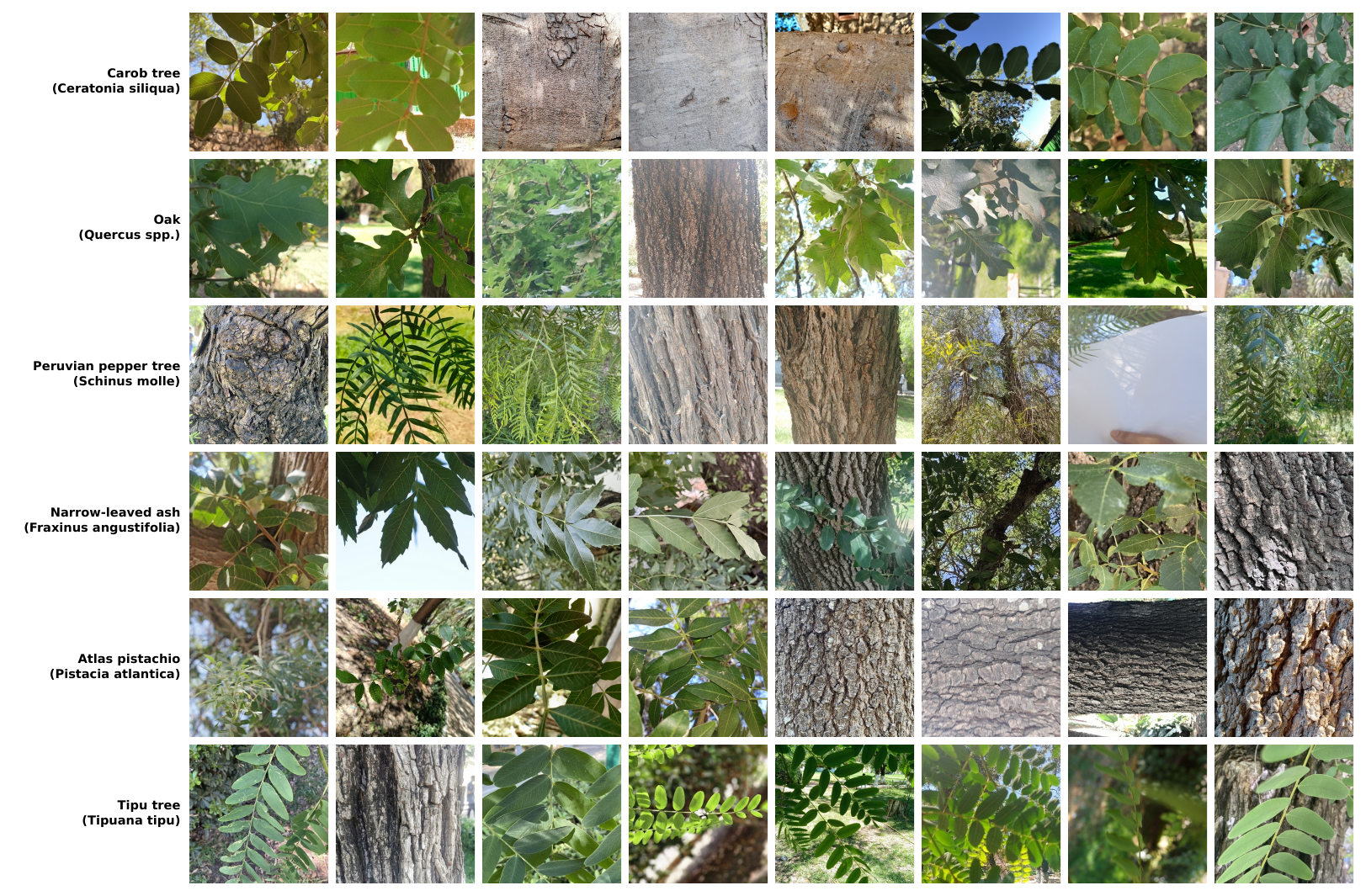}
    \caption{Samples from the AgrI Challenge dataset across the six target tree species. }
    \label{fig:dataset_samples}
\end{figure}

Figure~\ref{fig:dataset_samples} illustrates representative samples for each class, highlighting the diversity of visual appearance across collection teams.
 
\subsection{Dataset Characteristics}
 
This subsection presents the quantitative properties of the AgrI Challenge dataset, including its size, composition, distribution across teams, and the diversity of acquisition devices.

\subsubsection{Dataset Statistics}

Table~\ref{tab:dataset_distribution} presents the distribution of images collected by each team. The complete dataset comprises 50,673 images collected by 12 teams (11 participating teams plus the organizing committee).

\begin{table}[!htbp]
\centering
\caption{Distribution of images across teams and tree species classes}
\label{tab:dataset_distribution}
\footnotesize
\setlength{\tabcolsep}{3.5pt}
\renewcommand{\arraystretch}{1.05}
\begin{tabular}{lrrrrrrr}
\toprule
\textbf{Team} &
\makecell{\textbf{Carob}\\\textbf{tree}} &
\textbf{Oak} &
\makecell{\textbf{Peruvian}\\\textbf{pepper tree}} &
\textbf{Ash} &
\makecell{\textbf{Pistachio}\\\textbf{tree}} &
\makecell{\textbf{Tipu}\\\textbf{tree}} &
\textbf{Total} \\
\midrule
AI-4o & 960 & 1625 & 1221 & 934 & 1111 & 1493 & 7344 \\
AiGro & 686 & 604 & 595 & 572 & 541 & 635 & 3633 \\
CACTUS & 400 & 597 & 694 & 803 & 755 & 552 & 3801 \\
CHAJARA & 610 & 549 & 428 & 308 & 262 & 383 & 2540 \\
GreenAI & 675 & 705 & 609 & 559 & 722 & 515 & 3785 \\
PLT & 1089 & 1131 & 1000 & 1297 & 973 & 1066 & 6556 \\
RUSTICUS & 565 & 625 & 505 & 500 & 555 & 696 & 3446 \\
SMART AGRICULTURES & 638 & 923 & 1694 & 1214 & 1080 & 772 & 6321 \\
Scorpions & 422 & 450 & 407 & 456 & 620 & 440 & 2795 \\
Condimenteum & 1048 & 1000 & 655 & 1074 & 606 & 1121 & 5504 \\
The Neural Ninjas & 234 & 262 & 256 & 236 & 293 & 361 & 1642 \\
Organization team & 552 & 657 & 727 & 392 & 368 & 610 & 3306 \\
\midrule
\textbf{Total} & \textbf{7879} & \textbf{9128} & \textbf{8791} & \textbf{8345} & \textbf{7886} & \textbf{8644} & \textbf{50673} \\
\bottomrule
\end{tabular}
\end{table}

Dataset sizes ranged from 1,642 images (The Neural Ninjas team) to 7,344 images (AI-4o team). The overall dataset is relatively balanced across species, with class sizes ranging from 7,879 to 9,128 images.

\subsubsection{Device Diversity}

Images were captured using over 40 different device models, reflecting substantial diversity in acquisition conditions.
The dataset includes approximately 11{,}000 images where device information was not recorded, 6{,}900 images from iPhone~11,
5{,}400 images from Oppo~Reno5, 3{,}100 images from Samsung~Galaxy~A54~5G, and 2{,}400 images from Oppo~Reno7, along with
numerous additional devices represented with smaller counts.

\begin{figure}[!htbp]
    \centering
    \includegraphics[width=0.85\linewidth]{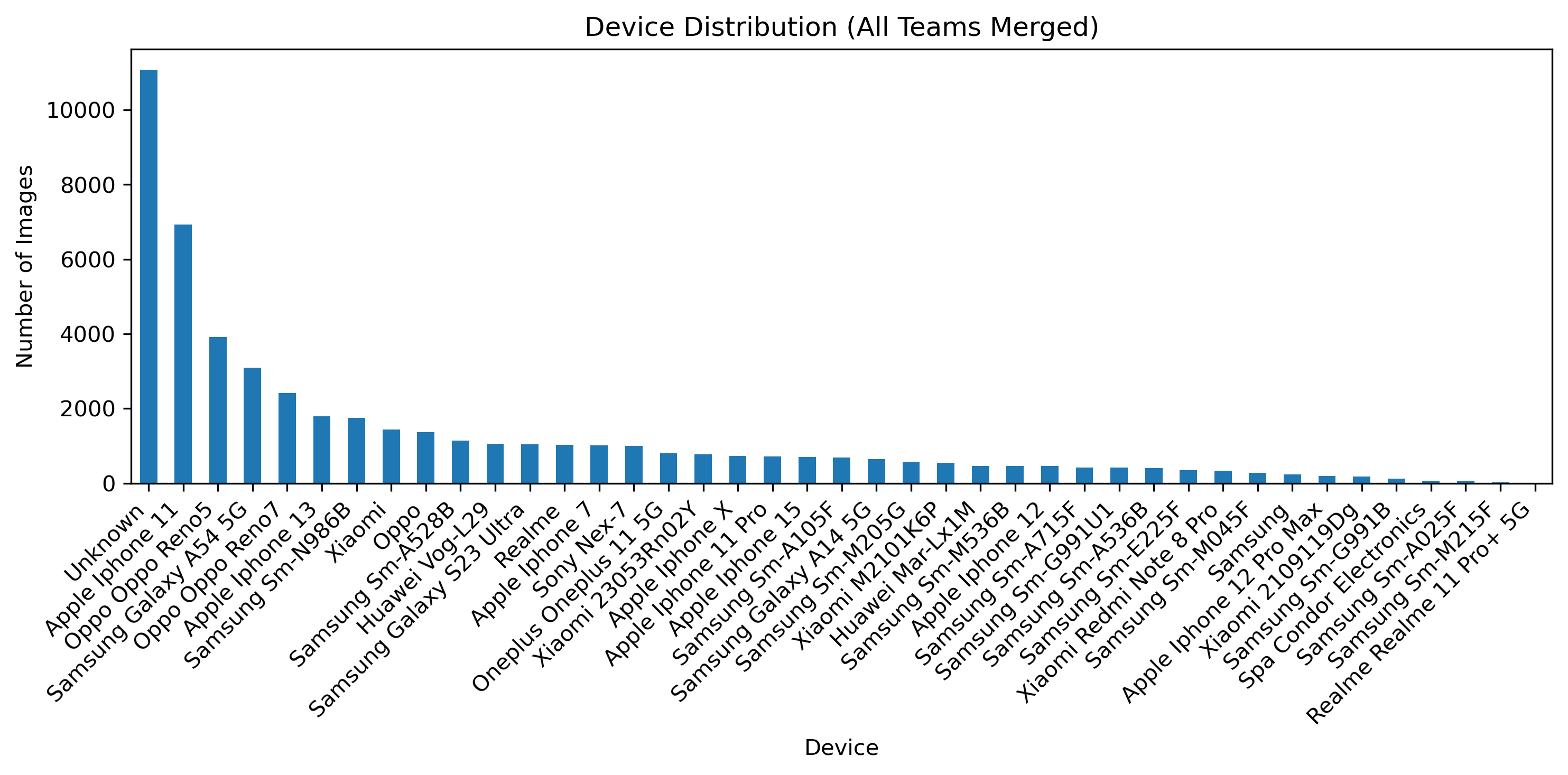}
    \caption{Histogram showing the distribution of dataset images by capture device model.
    The dataset contains images from more than 40 device types, with the largest groups originating
    from iPhone~11 and Oppo~Reno5, and a substantial portion labeled as Unknown due to missing metadata.}
    \label{fig:device_diversity_hist}
\end{figure}

The distribution of images across the most common capture devices is illustrated in Fig.~\ref{fig:device_diversity_hist},
highlighting both the variety of devices and the large subset of images lacking metadata.

\subsection{Dataset Preparation Pipeline}
This subsection describes the data preparation pipeline used to transform the raw, participant-collected images into a consistent and reliable dataset suitable for experimentation.

\subsubsection{Dataset Curation and Quality Assurance}

All images were systematically catalogued with standardized metadata, including file format, spatial resolution, file size, acquisition device (from EXIF), and a 64-bit perceptual hash (pHash) \citep{zauner2010implementation}. These attributes enable duplicate detection, integrity verification, and characterization of sensor and format variability, supporting reproducible analysis and facilitating future studies.

The dataset included seven common image formats (JPEG, PNG, HEIC/HEIF, BMP, TIFF), reflecting typical field acquisition diversity. Of the 50{,}673 images catalogued, JPEG predominated (75.6\%) followed by HEIC/HEIF (13.8\%), with remaining formats making up the rest. Two unreadable files ($\leq$ 0.01\%) were excluded. All metadata were preserved alongside the images to allow subsequent analysis, reuse, and benchmarking by other researchers.

\subsubsection{Duplicate Detection and Removal}

Cross-team duplicate detection was performed by grouping all images by their perceptual hash. A total of 13,579 records were involved in 6,370 duplicate groups, with group sizes ranging from 2 to 7 identical images. Duplicates arose primarily from images shared between teams or captured under identical conditions. Additionally, some duplicates originated when teams attempted to artificially increase the size of their datasets by using videos and extracting multiple frames, or by assembling collections of photos that are perceptually very similar; such images offer limited additional information and do not significantly contribute to training. Within each duplicate group, a single representative image was retained according to a deterministic priority rule: (i) largest file size, (ii) highest pixel resolution if tied, (iii) availability of device EXIF metadata, and (iv) team identity as a final tiebreaker. Applying this procedure, 7,209 duplicate images were removed, recovering approximately 9.6~GB of storage and ensuring that no image appeared in more than one team's data partition, a prerequisite for valid CTV evaluation.

\subsubsection{Resizing and Normalisation}

All retained images were resized to a uniform $336 \times 336$ pixel resolution. For square images, bicubic interpolation was applied directly. For non-square images, the shorter side was first scaled to 336 pixels while preserving the aspect ratio, and the result was then centre-cropped to the target size. Images smaller than the target size in either dimension were upscaled before cropping. All outputs were saved as JPEG with a quality factor of 95 to balance storage efficiency and visual fidelity. Class folder names were normalised to a common vocabulary (e.g., \texttt{chenes} $\rightarrow$ \texttt{chene}, \texttt{frenes} $\rightarrow$ \texttt{frene}). 


The final clean dataset comprised 47,367 images from 11 teams across 6 classes. Original images had a mean resolution of $2{,}709 \times 2{,}738$ pixels (range: $250$--$8{,}992$ px), and the resizing step reduced total storage from 121.7~GB to 2.9~GB (97.6\% reduction). The organisation team's data (3,306 images) was included in the CTV evaluation pool but not in the per-team statistics reported in Section~\ref{sec:results} to maintain consistency with the competition protocol.

\subsection{Cross-Team Validation Framework}
To evaluate model performance under realistic data variability, we adopt a Cross-Team Validation (CTV) evaluation paradigm, in which datasets collected by different teams are treated as distinct domains for training and testing. Unlike traditional cross-validation based on random data splits, CTV preserves the natural domain boundaries created by independent data collection processes. This setup better reflects real deployment scenarios where models must generalize to data originating from unseen teams, devices, or environments. Within this framework, we implement two complementary evaluation protocols: Train-on-One-Team-Only (TOTO) and Leave-One-Team-Out (LOTO), which are described in the following subsections.

For $n$ teams, CTV can be configured with various training ($S$) and testing ($T$) splits where $T \cap S = \emptyset$. This study implements two protocols.

\subsubsection{TOTO: Train-on-One-Team-Only Protocol}

The Train-on-One-Team-Only (TOTO) protocol evaluates cross-team generalization when models are trained using data collected by a single team. Because each team follows its own acquisition strategy (devices, viewpoints, environmental conditions), each dataset effectively represents a distinct domain. Training on one team and evaluating on the others therefore measures how well a model transfers to unseen data distributions.

For each team $i \in \{1,2,\ldots,n\}$:

\begin{itemize}
    \item \textbf{Training set}: 70\% of the data collected by team $i$.
    \item \textbf{Validation set}: The remaining 30\% of team $i$'s data.
    \item \textbf{Test sets}: All data collected by the other teams $j \neq i$.
\end{itemize}

The procedure is repeated for all teams, producing $n$ training runs and a complete cross-team evaluation matrix.

\subsubsection{LOTO: Leave-One-Team-Out Protocol}

The Leave-One-Team-Out (LOTO) protocol evaluates whether training on aggregated multi-source data improves robustness to domain shifts. In each fold, one team is held out for testing while the remaining teams form the training pool. Because the model is trained on the combined distribution of all other teams, the performance on the held-out team also provides an indirect indication of that team’s data quality and its consistency with the overall dataset. Large performance drops suggest that the held-out team’s data distribution deviates from the others.

For each team $i \in \{1,2,\ldots,n\}$:

\begin{itemize}
    \item \textbf{Training set}: Combined data from all teams except team $i$ (70\%).
    \item \textbf{Validation set}: 30\% of the training pool, stratified by team and class.
    \item \textbf{Test set}: All data from the held-out team $i$.
\end{itemize}

This process is repeated for every team, yielding $n$ evaluation folds in which each team serves once as the unseen test domain.

\subsection{Baseline Model Architectures}
To assess the effectiveness of the proposed framework, we compare its performance against representative baseline architectures from two dominant paradigms in computer vision: convolutional neural networks (CNNs) and transformer-based models. CNN architectures remain widely used in agricultural image analysis due to their efficiency in capturing local spatial patterns, while vision transformers have recently demonstrated strong capabilities in modeling long-range dependencies through attention mechanisms. Including both types of architectures provides a comprehensive benchmark for evaluating the proposed method. The baseline architectures considered in this study are summarized in Table~\ref{tab:baseline_architectures}.

\begin{table}[htbp]
\centering
\caption{Baseline model architectures used for comparison.}
\label{tab:baseline_architectures}
\begin{tabularx}{\textwidth}{l l c X}
\toprule
\textbf{Model} & \textbf{Type} & \textbf{Parameters} & \textbf{Purpose} \\
\midrule
DenseNet121 & CNN & 8M & Efficient backbone suitable for resource-constrained scenarios and strong feature reuse. \\
Swin-Tiny & Transformer & 28M & Lightweight transformer baseline capturing both local and global contextual information through window-based attention. \\
\bottomrule
\end{tabularx}
\end{table}

\textbf{DenseNet121.}
DenseNet121 \citep{huang2017densely} is a convolutional neural network characterized by dense connectivity between layers, where each layer receives feature maps from all preceding layers within the same dense block. This design encourages feature reuse, improves gradient propagation, and reduces the number of parameters compared to conventional deep CNN architectures. Owing to its computational efficiency and strong representational capability, DenseNet121 is widely adopted in image classification tasks, particularly in scenarios with limited computational resources.

\textbf{Swin Transformer (Tiny).}
The Swin Transformer \citep{liu2021swin} represents a hierarchical vision transformer architecture that introduces shifted window-based multi-head self-attention. Instead of computing global self-attention across the entire image, attention is restricted to local non-overlapping windows, significantly reducing computational complexity while maintaining strong modeling capacity. The shifted window mechanism enables cross-window interactions between consecutive layers, allowing the network to capture both local and global contextual information. The Swin-Tiny variant offers a lightweight transformer baseline with competitive performance and moderate computational requirements.

\subsection{Training Configuration}

All experiments were conducted on a shared GPU cluster equipped with NVIDIA H100 accelerators. GPU resources were partitioned across multiple users, and each training job was allocated approximately 22\,GB of GPU memory. Both baseline models were initialized with ImageNet-1K pre-trained weights to accelerate convergence and improve generalization. The models were optimized using the AdamW optimizer with weight decay, while the cross-entropy loss was used as the objective function. The learning rate followed a cosine annealing schedule to ensure stable convergence during training.

To ensure fair comparison, both architectures were trained using identical hyperparameters. The input images were resized to $224 \times 224$ pixels, and a dropout layer was applied in the classification head to mitigate overfitting. The main training hyperparameters are summarized in Table~\ref{tab:training_config}.

\begin{table}[htbp]
\centering
\caption{Training configuration used for all experiments.}
\label{tab:training_config}
\begin{tabular}{ll}
\toprule
\textbf{Parameter} & \textbf{Value} \\
\midrule
Hardware & NVIDIA H100 NVL (22 GB) \\
Pre-training & ImageNet-1K \\
Optimizer & AdamW \\
Weight decay & $10^{-4}$ \\
Initial learning rate & $10^{-4}$ \\
Learning rate schedule & Cosine annealing ($\eta_{min}=10^{-6}$) \\
Batch size & 32 \\
Epochs & 20 \\
Input resolution & $224 \times 224$ pixels \\
Dropout rate & 0.3 \\
Random seed & 42 \\
\bottomrule
\end{tabular}
\end{table}

\subsection{Evaluation Metrics}

Model performance is evaluated using classification accuracy and the validation--test gap (VTG). Accuracy measures the proportion of correctly classified samples. Given $N$ samples, with ground-truth label $y_i$ and predicted label $\hat{y}_i$, accuracy is defined as

\begin{equation}
\text{Accuracy} = \frac{1}{N} \sum_{i=1}^{N} \mathbf{1}(y_i = \hat{y}_i),
\end{equation}

where $\mathbf{1}(\cdot)$ is the indicator function. Since the dataset exhibits a relatively balanced distribution across the six species classes, accuracy provides a reliable measure of overall model performance.

To assess generalization, we also report the validation--test gap (VTG), defined as the difference between validation accuracy and test accuracy:

\begin{equation}
\text{VTG} = A_{\text{val}} - A_{\text{test}},
\end{equation}

where $A_{\text{val}}$ and $A_{\text{test}}$ denote validation and test accuracies, respectively. Smaller VTG values indicate better generalization.

\FloatBarrier

\section{Results}
\label{sec:results}
This section presents the experimental results obtained using the proposed evaluation framework. Model performance is examined under the two evaluation protocols described earlier: Single-Team Training (TOTO) and Leave-One-Team-Out (LOTO). Together, these experiments evaluate how training on single-source versus multi-source data affects cross-team generalization and model robustness.

\subsection{Single-Team Training (TOTO)}
This subsection analyzes model performance under the Single-Team Training (TOTO) protocol, where each model is trained on data from a single team and evaluated on datasets from the other teams. The analysis is organized into several parts: first, Controlled Evaluation with Standardized Architectures examines how baseline CNN and transformer models perform under uniform training conditions; next, Validation–Test Generalization Gaps quantifies the discrepancies between validation and test accuracy to highlight the impact of domain shifts; and finally, Per-Team Performance and Variability presents results broken down by individual held-out teams, revealing which teams’ data are easier or harder to generalize to. Together, these sections provide a detailed view of cross-team generalization in single-team training scenarios.

\subsubsection{Controlled Evaluation with Standardized Architectures}

Table~\ref{tab:toto_architecture_comparison} reports the TOTO results obtained using the standardized DenseNet121 and Swin Transformer architectures. Using identical training configurations ensures that performance differences across teams reflect variations in the collected datasets rather than model design.

\begin{table}[!htbp]
\centering
\caption{Controlled TOTO evaluation with standardized architectures}
\label{tab:toto_architecture_comparison}
\small
\begin{tabular}{@{}lrrrrrr@{}}
\toprule
\multirow{2}{*}{\textbf{Team}} & \multicolumn{3}{c}{\textbf{DenseNet121}} & \multicolumn{3}{c}{\textbf{Swin Transformer}} \\
\cmidrule(lr){2-4} \cmidrule(lr){5-7}
& \textbf{Val Acc.} & \textbf{Test Acc.} & \textbf{VTG} & \textbf{Val Acc.} & \textbf{Test Acc.} & \textbf{VTG} \\
\midrule
AI-4o & 96.60 & 86.94 & 9.65 & 98.09 & 91.08 & 7.02 \\
AiGro & 98.99 & 82.32 & 16.67 & 99.27 & 87.68 & 11.58 \\
CACTUS & 96.84 & 84.36 & 12.49 & 98.86 & 90.19 & 8.67 \\
CHAJARA & 97.51 & 81.56 & 15.94 & 98.82 & 88.03 & 10.79 \\
GreenAI & 96.92 & 85.47 & 11.45 & 98.42 & 90.86 & 7.55 \\
PLT & 98.53 & 75.50 & 23.03 & 99.90 & 84.00 & 15.90 \\
RUSTICUS & 97.87 & 83.58 & 14.30 & 98.26 & 86.96 & 11.30 \\
SMART AGRICULTURES & 97.79 & 87.83 & 9.95 & 98.95 & 93.02 & 5.93 \\
Scorpions & 97.62 & 82.85 & 14.76 & 98.81 & 89.57 & 9.23 \\
Condimenteum & 93.95 & 80.67 & 13.27 & 95.10 & 85.89 & 9.21 \\
The Neural Ninjas & 97.57 & 74.92 & 22.64 & 98.99 & 84.14 & 14.85 \\
Organization team & 98.59 & 68.32 & 30.27 & 99.60 & 75.16 & 24.44 \\
\midrule
\textbf{Mean} & \textbf{97.40} & \textbf{81.19} & \textbf{16.20} & \textbf{98.59} & \textbf{87.21} & \textbf{11.37} \\
\textbf{Std} & \textbf{1.25} & \textbf{5.42} & \textbf{6.07} & \textbf{1.16} & \textbf{4.51} & \textbf{5.24} \\
\bottomrule
\end{tabular}
\end{table}

Both architectures achieved high validation accuracy across all teams, with mean values of 97.40\% for DenseNet121 and 98.59\% for Swin Transformer. However, cross-team test accuracy was consistently lower, with mean values of 81.19\% and 87.21\%, respectively. Correspondingly, the average validation–test gap (VTG) reached 16.20\% for DenseNet121 and 11.37\% for Swin Transformer.

Across teams, the Swin Transformer consistently achieved higher test accuracy and lower VTG values than DenseNet121. Despite these architectural differences, team rankings remained highly consistent between the two models (Spearman $\rho = 0.94$), indicating that the relative ordering of team performance was largely preserved across architectures and reflecting similar generalization trends across datasets.

\begin{figure}[!htbp]
    \centering
    \includegraphics[width=\textwidth]{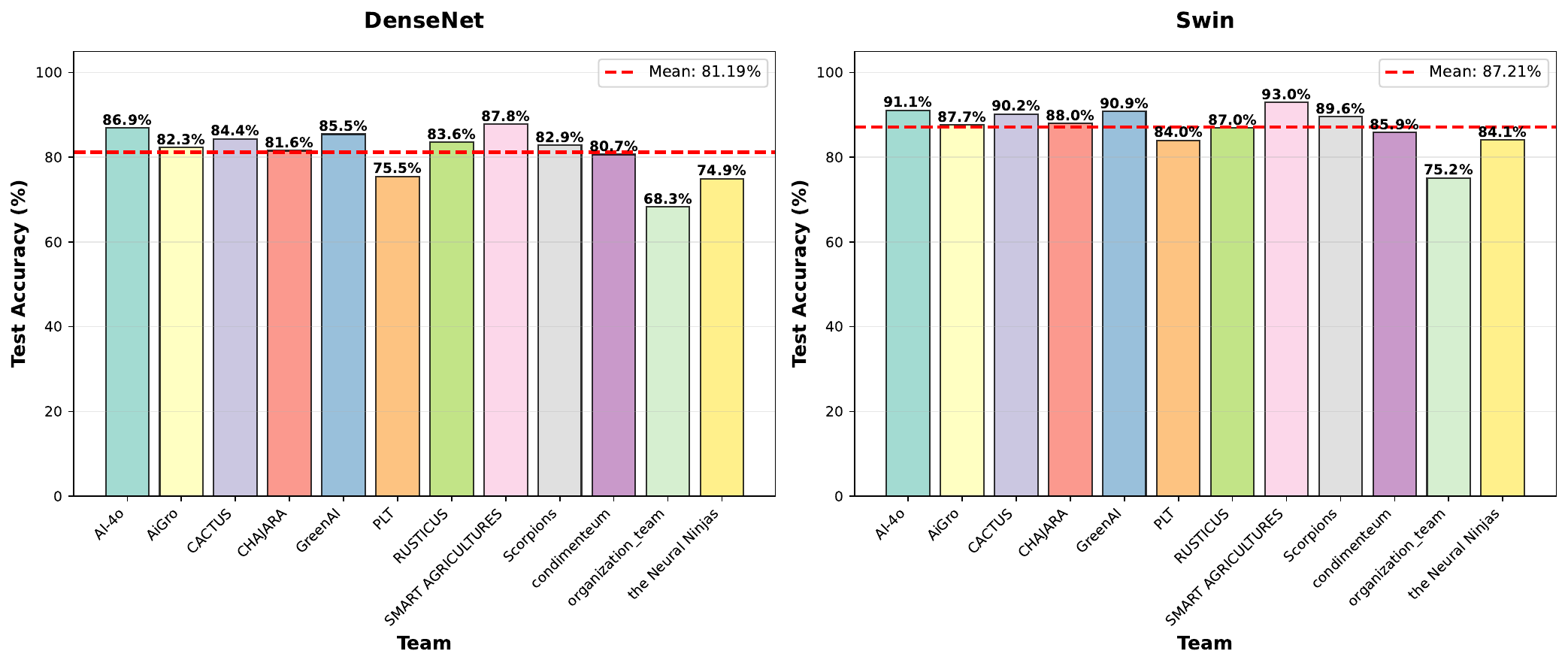}
    \caption{Cross-team test accuracy for DenseNet121 and Swin Transformer in TOTO protocol. Horizontal dashed lines indicate means: 81.19\% (DenseNet) and 87.21\% (Swin).}
    \label{fig:toto_comparison}
\end{figure}

Figure~\ref{fig:toto_comparison} visualizes the cross-team test accuracy for both architectures.

\subsubsection{Global Performance and Training Dynamics}

\begin{figure}[!htbp]
    \centering
    \includegraphics[width=\textwidth]{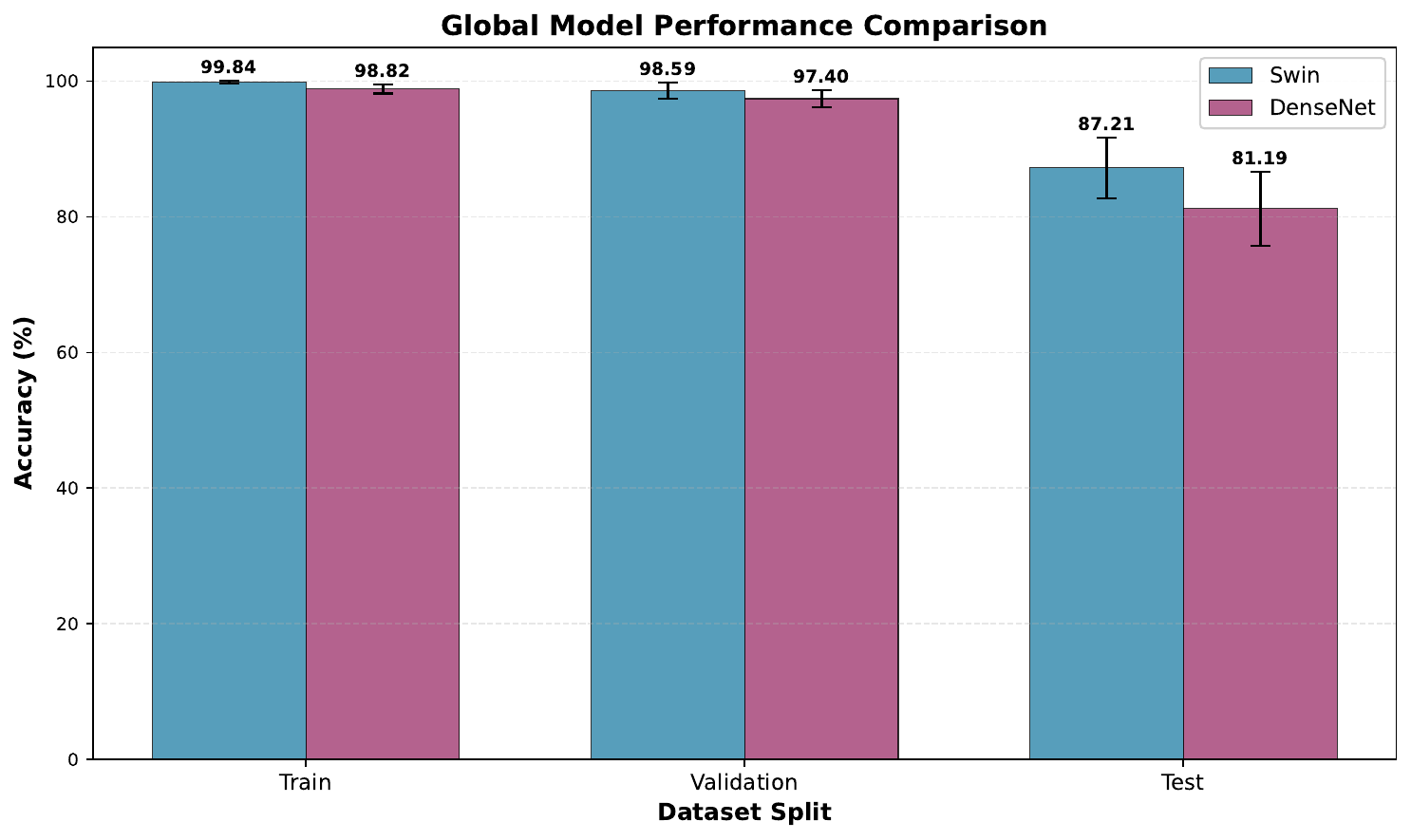}
    \caption{Global model performance under the TOTO protocol for DenseNet121 and Swin Transformer (mean $\pm$ std across 12 runs). The large validation--test gap highlights the cross-team generalization challenge under single-team training.}
    \label{fig:toto_global}
\end{figure}

Before examining epoch-wise learning behaviour, we summarize the overall performance across the train, validation, and cross-team test splits under the TOTO protocol. As shown in Figure~\ref{fig:toto_global}, Swin Transformer achieved mean training, validation, and test accuracies of 99.8\%, 98.6\%, and 87.2\%, respectively. DenseNet121 obtained accuracies of 98.8\% on training, 97.4\% on validation, and 81.2\% on cross-team test. The substantial separation between validation and test performance reflects the domain shift introduced when models trained on a single team's dataset are evaluated on independently collected datasets from other teams.

\begin{figure}[!htbp]
    \centering
    \includegraphics[width=\textwidth]{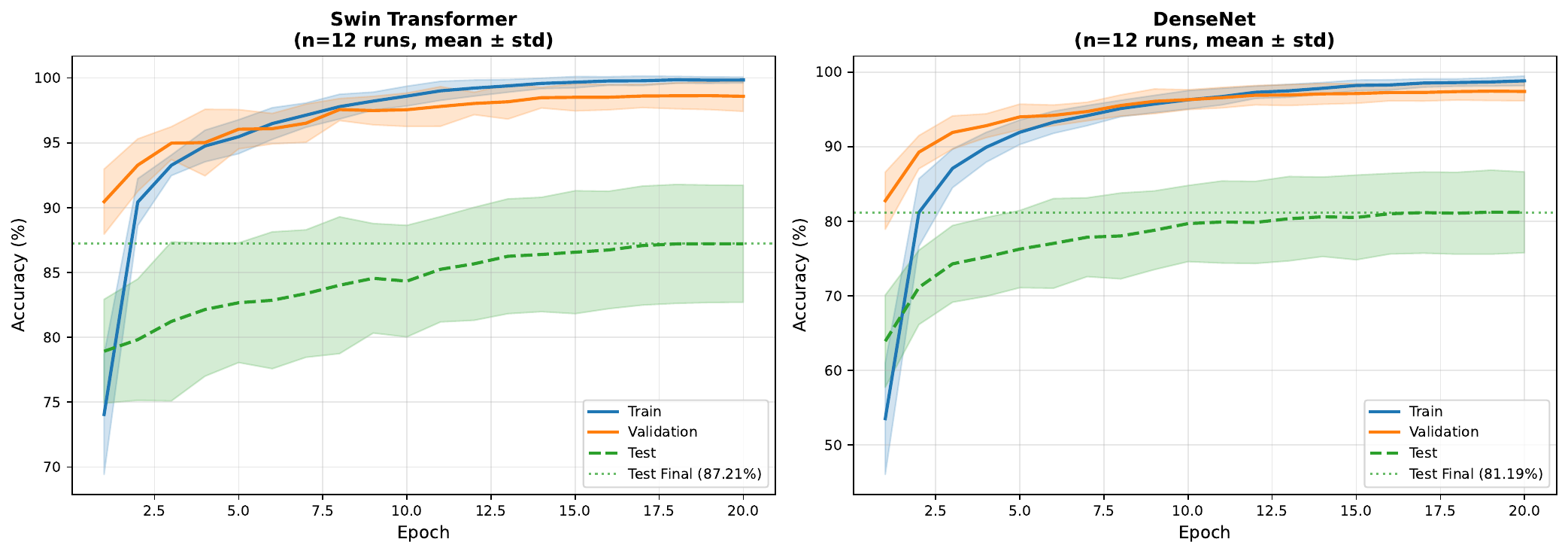}
    \caption{Aggregate learning curves for TOTO protocol ($n=12$ runs, mean $\pm$ std). Left: Swin Transformer. Right: DenseNet121. The gap between validation and test accuracy persists across all epochs.}
    \label{fig:learning_curves}
\end{figure}

Figure~\ref{fig:learning_curves} presents aggregate learning curves across all 12 TOTO runs. For DenseNet121, training accuracy reached near 100\%, validation stabilized around 97\%, and test plateaued at 81.19\%. For Swin Transformer, training reached near 100\%, validation stabilized around 98\%, and test plateaued at 87.21\%.

The gap between validation and test accuracy remained stable throughout all training epochs. The shaded regions indicate standard deviation across runs, with DenseNet showing larger variance in test accuracy.

\subsubsection{Validation--Test Gap Analysis}

\begin{figure}[!htbp]
    \centering
    \includegraphics[width=0.75\textwidth]{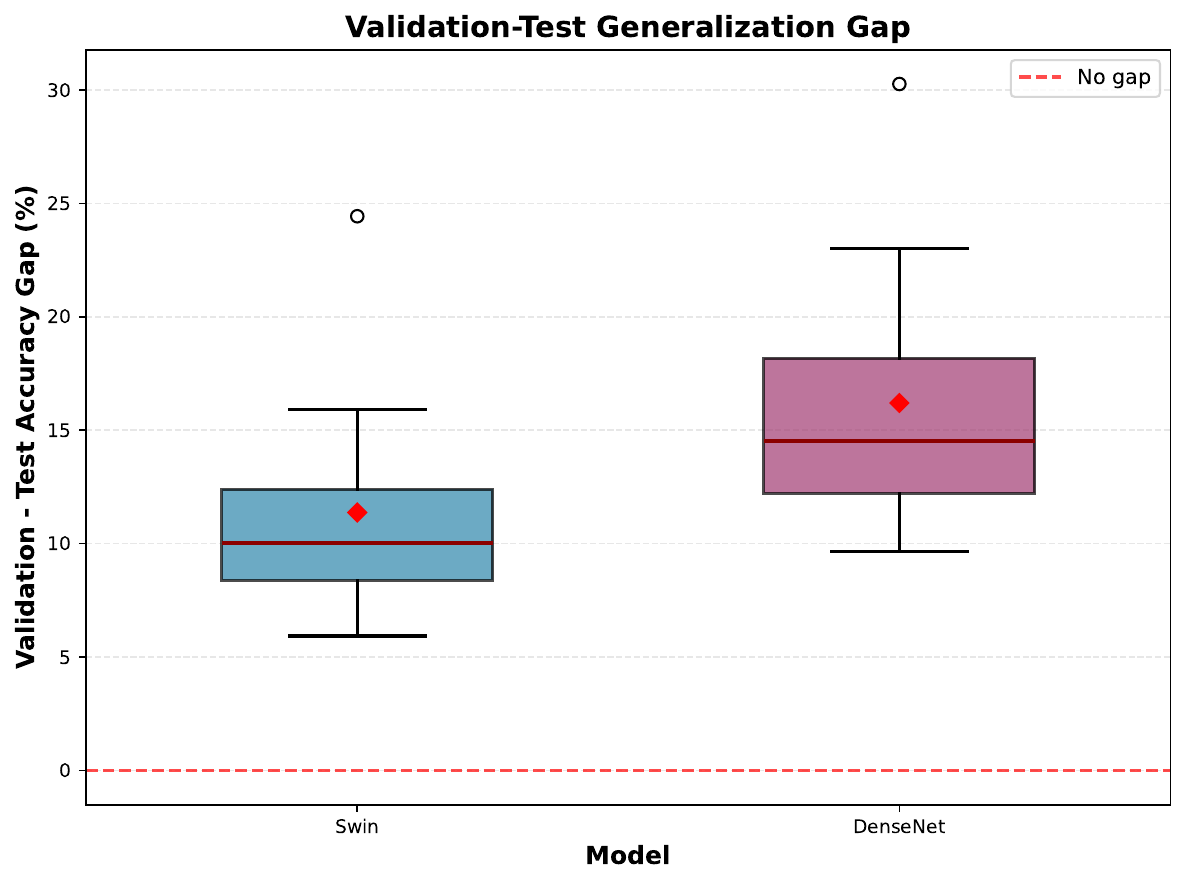}
    \caption{Distribution of validation--test gaps under the TOTO protocol for
    Swin Transformer and DenseNet121. The red dashed line marks zero gap.
    Larger gaps indicate stronger cross-team generalization loss.}
    \label{fig:toto_gap}
\end{figure}

Figure~\ref{fig:toto_gap} summarizes the distribution of validation--test gaps (VTG)
across teams for the two architectures under the TOTO protocol. The VTG measures the
difference between in-distribution validation accuracy and cross-team test accuracy,
providing a direct view of the generalization gap.

Across all runs, both models exhibit positive gaps, confirming that validation
performance consistently exceeds cross-team test performance. DenseNet121 shows
larger and more variable gaps, while Swin Transformer exhibits a narrower
distribution with generally smaller VTG values. These observations are consistent
with the aggregate statistics reported in Table~\ref{tab:toto_architecture_comparison}.

\subsubsection{Cross-Team Compatibility}

\begin{figure}[!htbp]
    \centering
    \includegraphics[width=\textwidth]{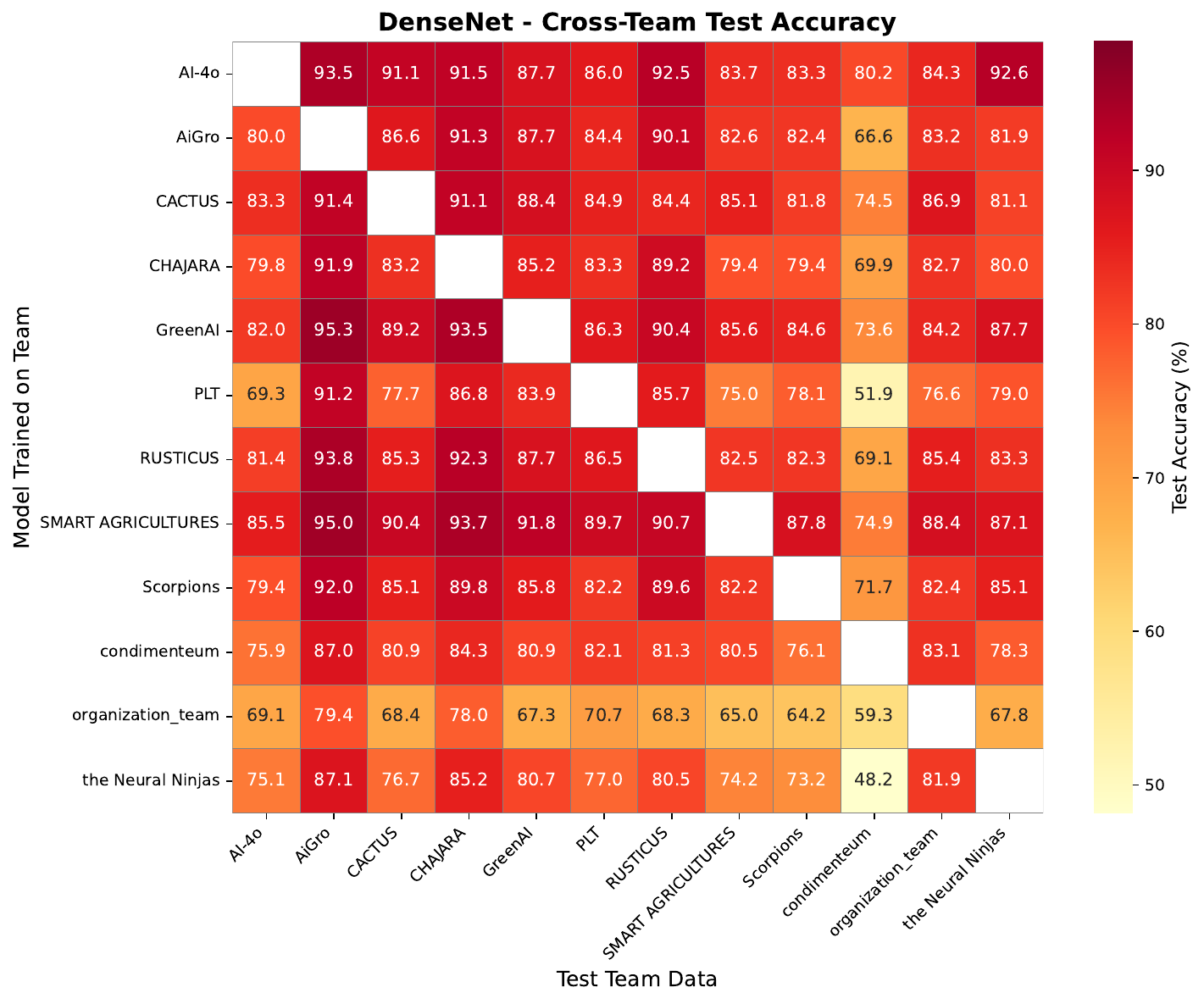}
    \caption{DenseNet121 cross-team accuracy matrix. Rows: training teams; columns: test teams. Range: 48.2\%--95.3\%.}
    \label{fig:densenet_heatmap}
\end{figure}

\begin{figure}[!htbp]
    \centering
    \includegraphics[width=\textwidth]{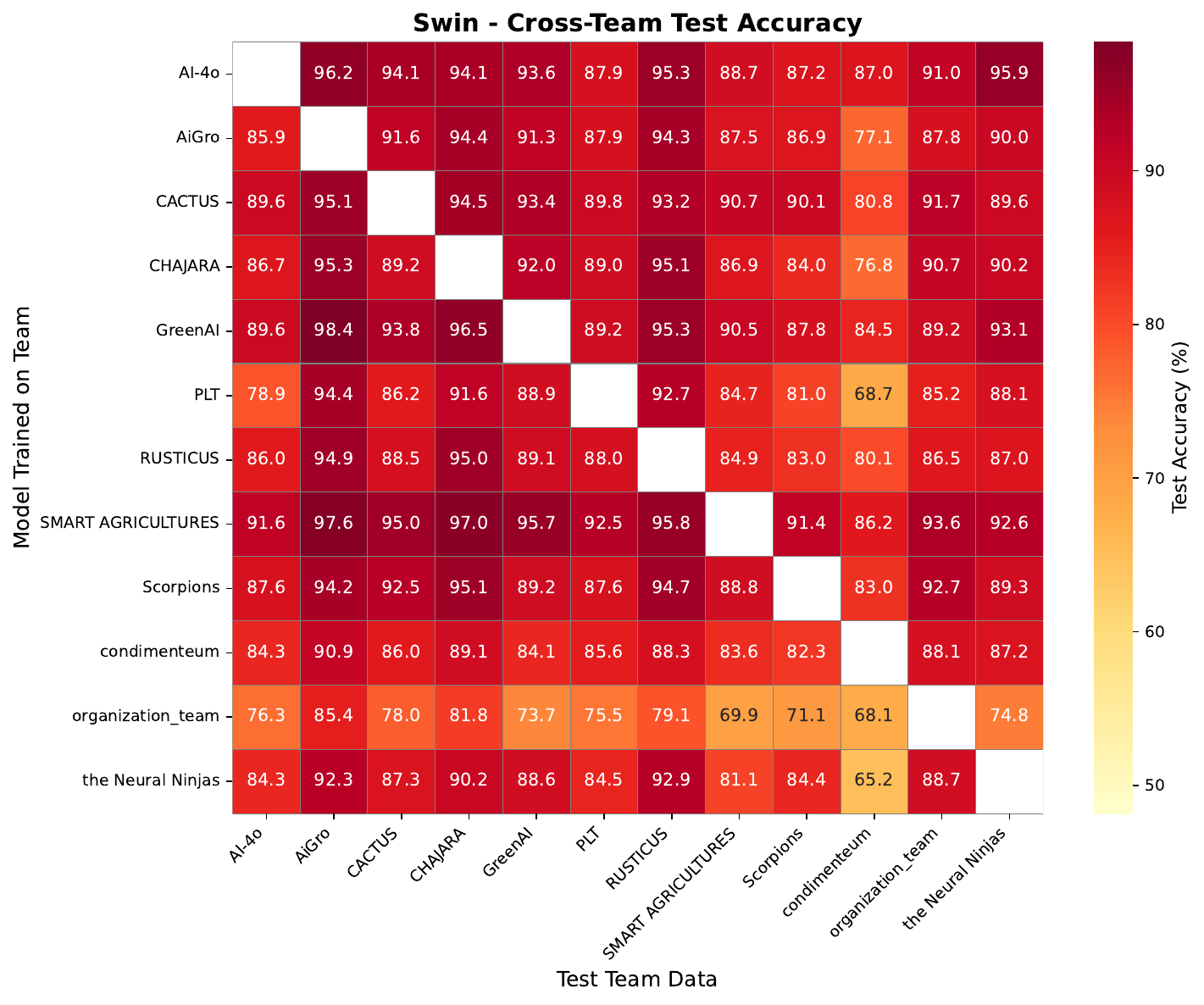}
    \caption{Swin Transformer cross-team accuracy matrix. Range: 65.2\%--98.4\%.}
    \label{fig:swin_heatmap}
\end{figure}

Figures~\ref{fig:densenet_heatmap} and~\ref{fig:swin_heatmap} visualize the cross-team
accuracy matrices obtained under the TOTO protocol. Each row corresponds to a model
trained on one team’s dataset, while each column represents evaluation on another
team’s data. These matrices therefore summarize the full pairwise transfer performance
between all teams.

For DenseNet121, cross-team accuracies ranged from 48.2\% to 95.3\%, whereas Swin
Transformer produced values between 65.2\% and 98.4\%. Despite the architectural
differences, the global patterns of the two matrices were strongly aligned, with a high
correlation between corresponding entries (Pearson $r = 0.95$\footnote{The correlation
was computed by flattening the two cross-team accuracy matrices into vectors and
calculating the Pearson correlation between their corresponding entries.}).

Row-wise averages provide a compact view of how well models trained on each team’s
dataset transfer to the other teams. For DenseNet, these averages ranged from 68.1\%
for the Organization team to 88.7\% for AI-4o. Column-wise averages capture the
relative difficulty of each dataset when used as the test domain, spanning from 74.0\%
for Condimenteum to 90.5\% for AiGro.

Overall, the matrices reveal substantial variability across training–testing combinations,
with consistent relative trends observed across the two architectures. The detailed
implications of these cross-team patterns are examined further in the discussion section.

\FloatBarrier

\subsection{Collaborative Multi-Team Training (LOTO)}
This subsection presents results under the Collaborative Multi-Team Training (LOTO) protocol, in which models are trained on aggregated data from all teams except one and evaluated on the held-out team. The analysis is organized into several parts: Validation–Test Generalization Gaps quantifies how well models trained on multi-source data generalize to unseen teams; Global Performance and Training Dynamics examines aggregate accuracy and the alignment between training, validation, and test distributions; and Per-Team Performance Analysis provides a breakdown of test accuracy for each held-out team, highlighting residual variability attributable to dataset-level properties rather than model architecture. Together, these analyses demonstrate the benefits of collaborative, multi-source training in improving cross-team generalization and robustness.

\subsubsection{Per-Team Results}

Table~\ref{tab:loto_architecture_comparison} present LOTO results for DenseNet121 and Swin Transformer respectively. In each fold, models were trained on data from 11 teams and evaluated on the single held-out team.

\begin{table}[!htbp]
\centering
\caption{LOTO evaluation using standardized DenseNet121 and Swin Transformer architectures.}
\label{tab:loto_architecture_comparison}
\small
\begin{tabular}{@{}lrrrrrr@{}}
\toprule
\multirow{2}{*}{\textbf{Held-Out Team}} & \multicolumn{3}{c}{\textbf{DenseNet121}} & \multicolumn{3}{c}{\textbf{Swin Transformer}} \\
\cmidrule(lr){2-4} \cmidrule(lr){5-7}
& \textbf{Val Acc.} & \textbf{Test Acc.} & \textbf{VTG} & \textbf{Val Acc.} & \textbf{Test Acc.} & \textbf{VTG} \\
\midrule
Condimenteum         & 98.37 & 87.61 & +10.76 & 99.03 & 92.57 & +6.46 \\
AI-4o                & 98.28 & 92.96 & +5.32  & 98.97 & 96.76 & +2.21 \\
Scorpions            & 98.11 & 93.35 & +4.77  & 98.93 & 95.46 & +3.47 \\
PLT                  & 97.91 & 93.50 & +4.41  & 98.61 & 94.78 & +3.83 \\
SMART AGRICULTURES   & 97.98 & 94.95 & +3.03  & 98.83 & 96.57 & +2.26 \\
GreenAI              & 98.04 & 97.17 & +0.86  & 98.77 & 97.86 & +0.91 \\
RUSTICUS             & 98.12 & 97.19 & +0.94  & 98.75 & 98.23 & +0.52 \\
The Neural Ninjas    & 98.15 & 97.69 & +0.47  & 98.80 & 98.66 & +0.14 \\
CACTUS               & 98.07 & 97.71 & +0.35  & 98.76 & 98.58 & +0.18 \\
AiGro                & 98.01 & 98.73 & $-$0.73 & 98.73 & 99.37 & $-$0.64 \\
CHAJARA              & 98.25 & 98.94 & $-$0.68 & 98.86 & 99.57 & $-$0.71 \\
Organization team    & 98.23 & 93.89 & +4.34  & 98.79 & 96.07 & +2.72 \\
\midrule
\textbf{Mean} & \textbf{98.13} & \textbf{95.31} & \textbf{+2.82} & \textbf{98.82} & \textbf{97.04} & \textbf{+1.78} \\
\textbf{Std}  & \textbf{0.14}  & \textbf{3.27}  & \textbf{3.33}  & \textbf{0.11}  & \textbf{2.08}  & \textbf{2.13} \\
\bottomrule
\end{tabular}
\end{table}

For DenseNet121, the mean test accuracy across all held-out teams was 95.31\% ($\pm$3.27\%), with performance ranging from 87.61\% (Condimenteum) to 98.94\% (CHAJARA). For the Swin Transformer, the mean test accuracy increased to 97.04\% ($\pm$2.08\%), with results spanning 92.57\% (Condimenteum) to 99.57\% (CHAJARA). Although the transformer-based model achieved higher overall accuracy and lower variability, the relative ordering of teams remained largely unchanged.

To quantify this consistency, the Pearson correlation coefficient ($r$) was computed between the test accuracies obtained by the two architectures across the 12 held-out teams. Pearson correlation measures the strength of the linear relationship between two sets of values; here, it captures whether teams that are difficult for one architecture also tend to be difficult for the other. The resulting value ($r = 0.97$) indicates a very strong correspondence between architectures. This suggests that teams that yield higher or lower performance for one model tend to produce similar outcomes for the other, implying that the observed performance differences are primarily driven by dataset characteristics associated with each team rather than by the choice of model architecture.

\subsubsection{Validation-Test Gap Under Collaborative Training}

\begin{figure}[!htbp]
    \centering
    \includegraphics[width=0.75\textwidth]{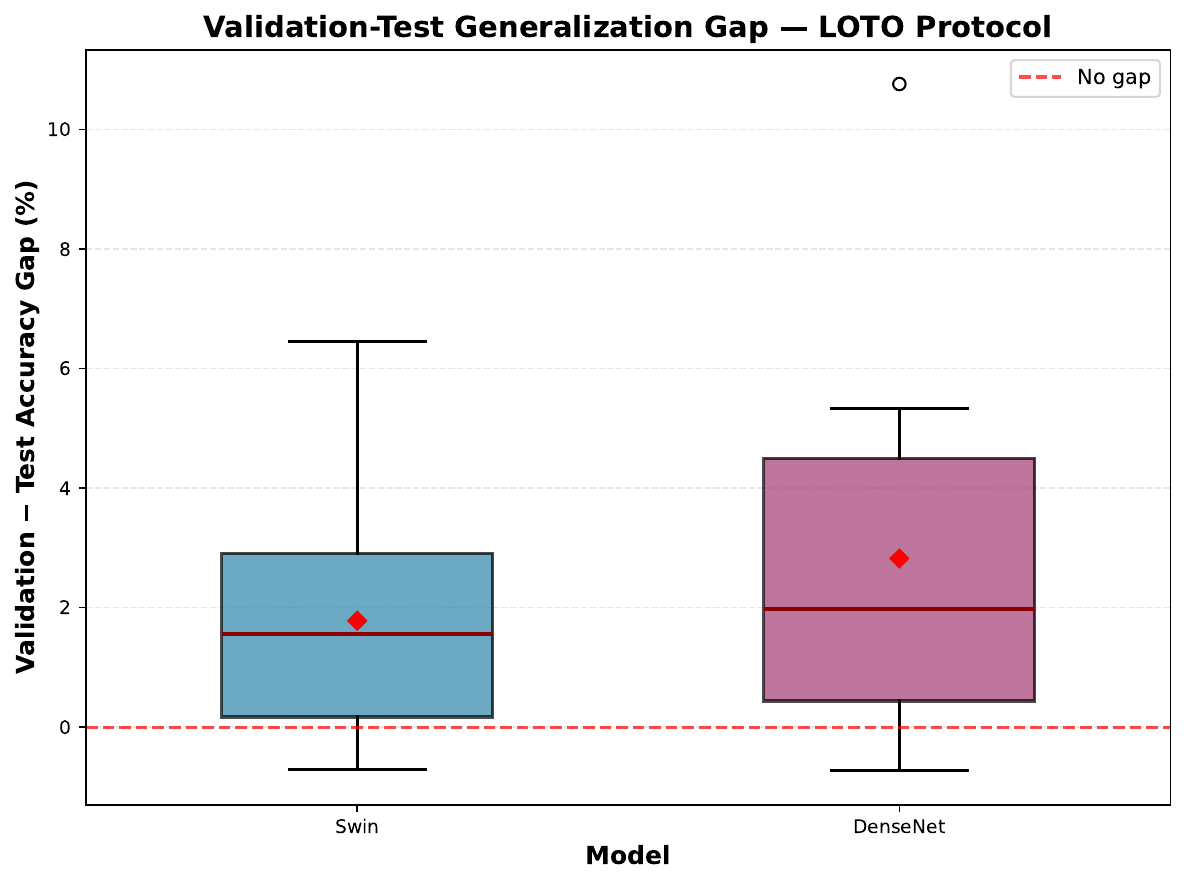}
    \caption{Distribution of validation-test gaps under LOTO protocol for Swin Transformer and DenseNet121. The red dashed line marks zero gap. Two teams achieved negative gaps in both models, indicating the held-out data was easier to generalize to than the validation set. The single outlier (DenseNet, Condimenteum: +10.76\%) reflects that team's distinctive collection methodology.}
    \label{fig:loto_gap}
\end{figure}

A striking result is the dramatic reduction of the validation-test gap under LOTO. The mean VTG dropped from 16.20\% (TOTO DenseNet) to 2.82\% (LOTO DenseNet), and from 11.37\% (TOTO Swin) to 1.78\% (LOTO Swin), reductions of 82\% and 84\% respectively. Figure~\ref{fig:loto_gap} illustrates the distribution of VTGs across teams for both architectures.

Notably, two teams (AiGro and CHAJARA) exhibited \emph{negative} VTGs under both architectures (e.g., AiGro: $-$0.73\% DenseNet, $-$0.64\% Swin; CHAJARA: $-$0.68\% DenseNet, $-$0.71\% Swin). A negative VTG indicates that the held-out team's test data was more representative of the multi-source training distribution than the heterogeneous validation set itself, suggesting that these teams employed highly consistent collection protocols aligned with the majority of the dataset.

\subsubsection{Global Performance and Training Dynamics}

\begin{figure}[!htbp]
    \centering
    \includegraphics[width=\textwidth]{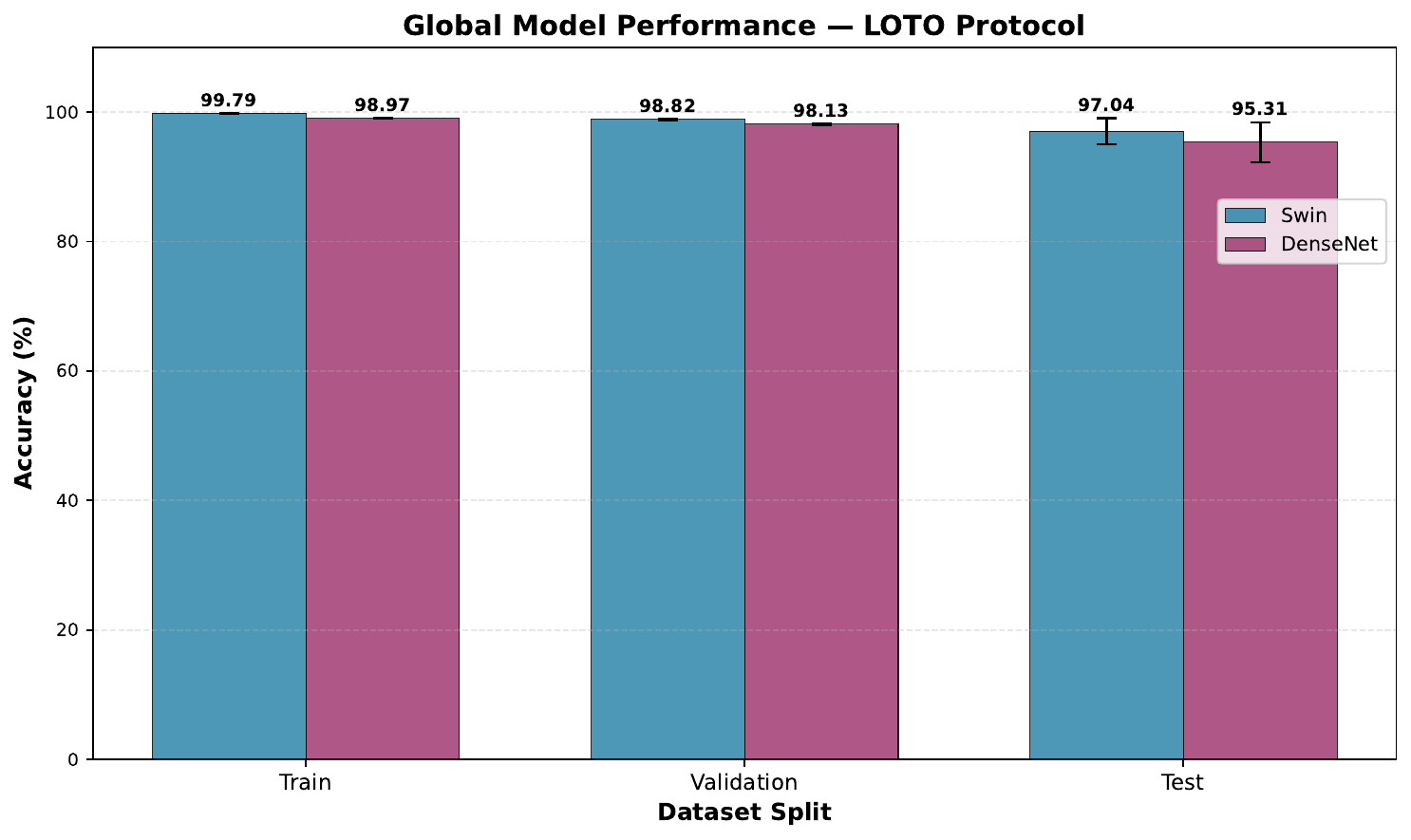}
    \caption{Global model performance under LOTO protocol for DenseNet121 and Swin Transformer (mean $\pm$ std across 12 folds). The compressed train-validation-test spread contrasts sharply with the large gaps observed under TOTO.}
    \label{fig:loto_global}
\end{figure}

Figure~\ref{fig:loto_global} shows aggregate global performance across all three splits. Under LOTO, training accuracy reached 99.79\% (Swin) and 98.97\% (DenseNet), validation accuracy 98.82\% and 98.13\%, and mean test accuracy 97.04\% and 95.31\%. The much narrower train-validation-test spread relative to TOTO confirms that multi-source training substantially aligns the learned representation with out-of-distribution data.

\begin{figure}[!htbp]
    \centering
    \includegraphics[width=\textwidth]{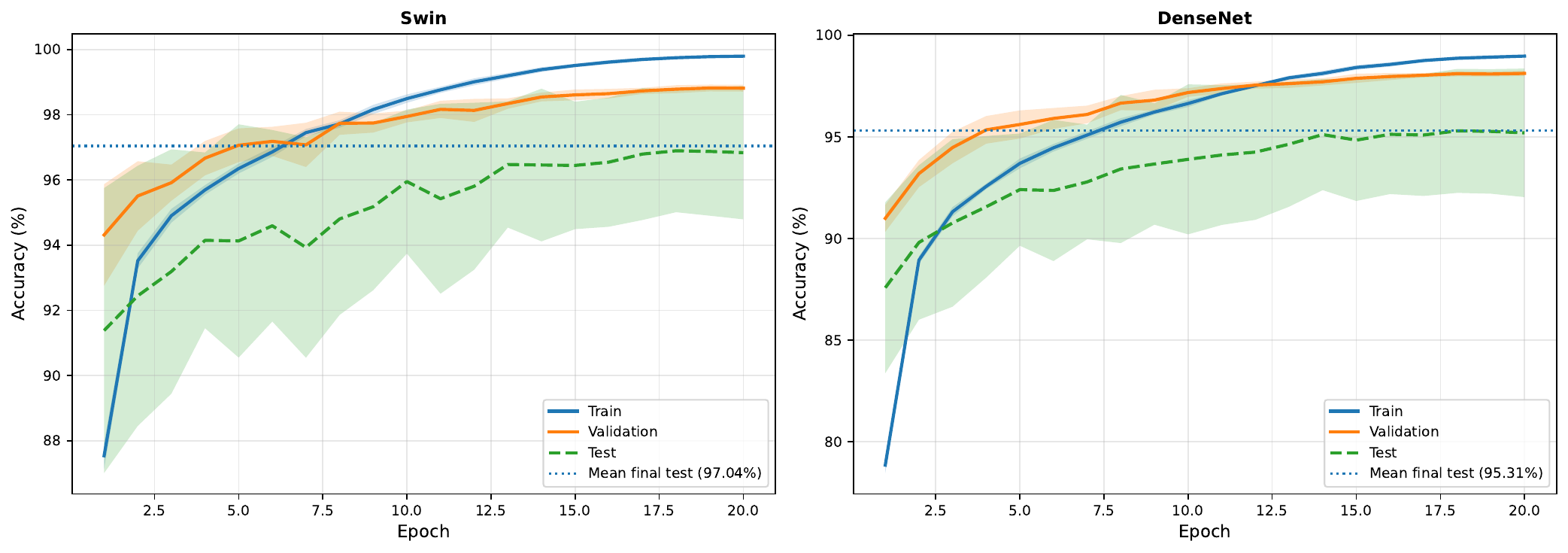}
    \caption{Aggregate LOTO learning curves. Left: Swin Transformer (mean final test: 97.04\%). Right: DenseNet121 (mean final test: 95.31\%).}
    \label{fig:loto_curves}
\end{figure}

Figure~\ref{fig:loto_curves} presents the aggregate learning curves across all 12 LOTO folds. Test accuracy rises steadily throughout training and converges towards the validation curve by epoch 20, in striking contrast to the TOTO curves in Figure~\ref{fig:learning_curves}.

\begin{figure}[!htbp]
    \centering
    \includegraphics[width=\textwidth]{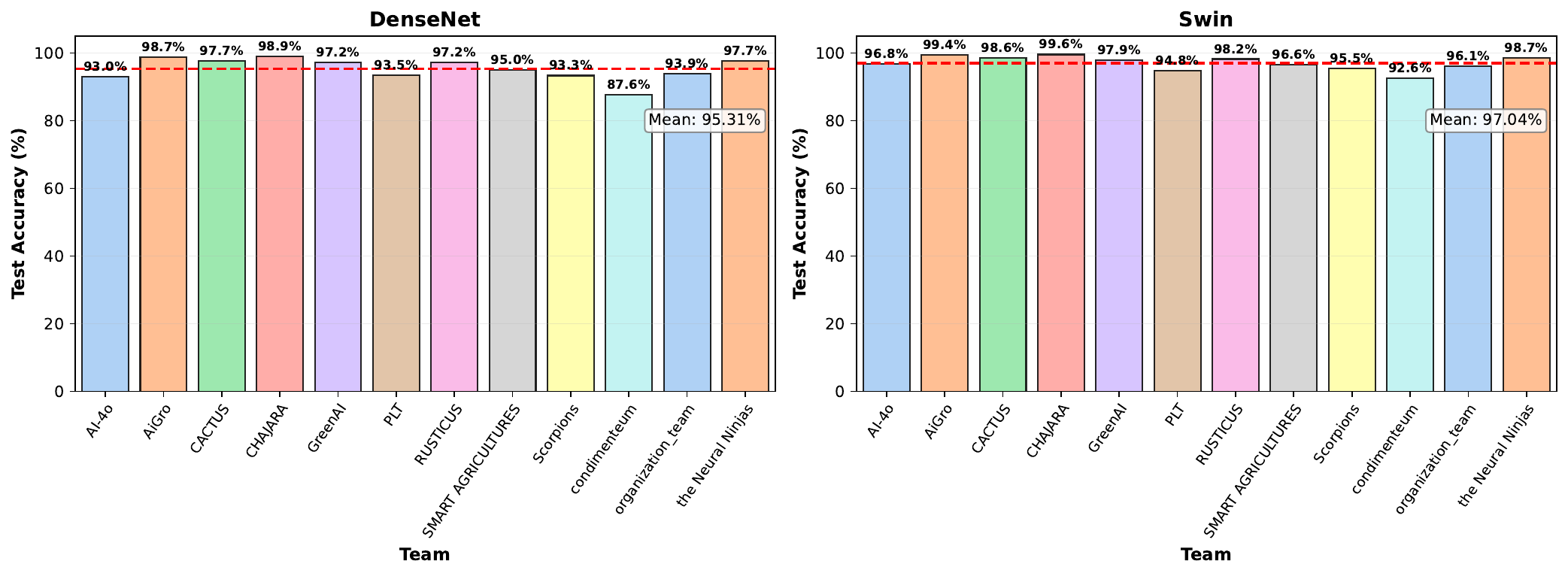}
    \caption{Per-held-out-team test accuracy under LOTO protocol for Swin Transformer (left) and DenseNet121 (right). Dashed blue lines indicate overall means (97.04\% and 95.31\%). The consistency of team rankings across both architectures (Spearman $\rho = 0.92$) confirms that residual variation reflects dataset-level properties rather than architecture-specific effects.}
    \label{fig:loto_perteam}
\end{figure}

Figure~\ref{fig:loto_perteam} provides a per-team breakdown, allowing direct comparison of which held-out teams benefit most and least from collaborative training.

\FloatBarrier

\section{Discussion}
This section interprets the experimental findings in the context of cross-team generalization and data-centric learning. We first examine the validation–test gaps under single-team training (TOTO) to highlight the limitations of models trained on isolated datasets. Next, we analyze the impact of collaborative multi-team training (LOTO) on overall performance, variance, and generalization, illustrating how multi-source data improves robustness. We then compare the performance of the two architectures to assess the relative influence of model choice versus dataset diversity. Finally, we investigate residual variation attributable to dataset characteristics and discuss limitations of the study, providing perspective on the broader applicability of the CTV framework and identifying avenues for future work.

\subsection{Validation-Test Gap Under Single-Team Training}

The TOTO results reveal a systematic and persistent gap between in-distribution validation performance and cross-team test performance. In controlled evaluation (Table~\ref{tab:toto_architecture_comparison}), mean VTGs reached 16.20\% (DenseNet) and 11.37\% (Swin). Figure~\ref{fig:learning_curves} shows a persistent gap, indicating that the models do not reduce this difference with additional training. This suggests that the gap is caused by distributional shift rather than a lack of sufficient training.

The severity of the gap varied considerably across teams. Organization team exhibited the largest gap (30.27\% DenseNet, 24.44\% Swin), indicating that its data was least representative of the multi-team distribution, a pattern that is consistent with the lowest TOTO test accuracy observed for that team (68.32\% DenseNet). Conversely, SMART AGRICULTURES achieved the smallest gap (9.95\% DenseNet, 5.93\% Swin) alongside the highest test accuracy, suggesting its sampling methodology produced more diverse, field-representative images.

\subsection{Collaborative Training Dramatically Improves Generalization}

The transition from TOTO to LOTO yields the most striking finding of this study. Pooling data from 11 teams improved mean test accuracy by 14.12 percentage points for DenseNet (81.19\% $\to$ 95.31\%) and 9.83 percentage points for Swin (87.21\% $\to$ 97.04\%). Equally important, performance variance was substantially reduced: standard deviation fell from 5.42\% to 3.27\% for DenseNet (a 40\% reduction) and from 4.51\% to 2.08\% for Swin (a 54\% reduction).

The validation-test gap reduction is even more dramatic: an 82\% reduction for DenseNet (16.20\% $\to$ 2.82\%) and an 84\% reduction for Swin (11.37\% $\to$ 1.78\%). This demonstrates that the gap under TOTO is almost entirely attributable to the narrow diversity of single-team training data, not to fundamental architectural limitations or overfitting.

The most striking individual case is the Organization team, which went from the lowest TOTO test accuracy (68.32\% DenseNet) to a competitive 93.89\% under LOTO, a gain of 25.57 percentage points. This team's dataset, while exhibiting distinctive collection characteristics that limited single-source generalization, benefited most from exposure to the diversity of the remaining 11 teams. The finding reinforces the core CTV thesis: datasets that appear problematic in isolation may be perfectly adequate as part of a diverse collaborative pool.

\subsection{Architecture Comparison}

Swin Transformer consistently outperformed DenseNet121 under both protocols, a 6.02 pp advantage in TOTO and a 1.73 pp advantage in LOTO. Notably, the advantage narrows under collaborative training, suggesting that data diversity compensates partially for the architectural gap. In both protocols, team rankings were highly consistent across architectures (TOTO: Spearman $\rho = 0.94$; LOTO: $\rho = 0.97$), confirming that performance differences are fundamentally driven by dataset characteristics rather than model choice. This result directly supports the data-centric AI premise: once sufficient data diversity is present, architecture choice has diminishing returns.

The validation accuracy was remarkably stable across all LOTO folds for both architectures (DenseNet: 98.13\% $\pm$ 0.14\%; Swin: 98.82\% $\pm$ 0.11\%), in contrast to test accuracy which varied by over 11 percentage points across held-out teams. This divergence confirms that validation accuracy, measured on a random split of the training pool, is not a reliable proxy for held-out generalization and highlights the necessity of CTV as an evaluation paradigm.

\subsection{Dataset Characteristics Drive Residual Variation} \label{sec:discussion_errors}

Under LOTO, Condimenteum consistently achieved the lowest test accuracy (87.61\% DenseNet, 92.57\% Swin) despite being held out from a pool of 11 diverse teams. Its validation-test gap remained the largest (10.76\% DenseNet), indicating that Condimenteum’s test distribution is genuinely different from the combined training distribution of the other 11 teams. In contrast, CHAJARA and AiGro achieved the highest test accuracies and negative VTGs, implying that their data distributions overlap substantially with the aggregate multi-team distribution.

Additional insight into these dataset-level differences can be obtained from the cross-team accuracy matrices presented in Figures~\ref{fig:densenet_heatmap} and~\ref{fig:swin_heatmap}. Because rows correspond to training teams and columns correspond to test teams, the heatmaps provide a complementary perspective on how datasets behave both as sources of training data and as unseen test domains.

Rows that remain consistently dark red indicate training datasets that generalize well across teams. For example, teams such as AI-4o and SMART AGRICULTURES exhibit strong performance across many test domains, suggesting that their collected data capture a broader range of visual conditions and environmental variability. In contrast, rows with systematically lighter colors indicate training datasets that produce weaker cross-domain generalization. The Organization team provides the clearest example of this pattern, as models trained solely on its data achieve relatively low accuracy across many test teams.

Columns reveal the complementary perspective: the difficulty of a dataset when used as the unseen test domain. Columns that remain strongly red across many training rows correspond to datasets that are relatively easy for models trained elsewhere to recognize. For example, the AiGro dataset shows consistently high cross-team accuracy across both architectures. Conversely, columns with lighter colors indicate more challenging domains that models struggle to generalize to. The Condimenteum column illustrates this behavior, showing systematically lower cross-team performance across both DenseNet121 and Swin Transformer.

Importantly, these row and column patterns appear consistently across the two architectures. This agreement indicates that the observed variability primarily reflects dataset properties rather than model-specific behavior.

These results indicate that cross-team generalization is strongly influenced by how distinctive a team’s collection distribution is relative to the pooled training data. Teams whose held-out data aligns more closely with the multi-team mixture achieve higher test accuracy and smaller (or negative) validation–test gaps, while teams with more distinctive collection practices remain harder to generalize to even under collaborative training.

\subsection{Limitations and Future Work}

This study focused on six tree species within a single institutional site (ENSA Botanical Garden, Algiers) during a two-day collection window. While the CTV framework itself is task- and domain-agnostic, the numerical findings---particularly the magnitude of validation--test gaps and the benefit of collaborative training---may not transfer directly to other species counts, crop types, climates, or phenological stages. The botanical garden setting, though providing verified taxonomic labels, also implies spatial proximity among specimens, meaning that background scenes, soil appearance, and lighting conditions are partially shared across classes. Field deployment in open agricultural landscapes would introduce greater background heterogeneity and potentially larger domain shifts than those observed here.

The visual features used for classification (Section~\ref{sec:visual_cues}) are restricted to vegetative structures, leaves, trunk, and canopy, as reproductive organs were excluded due to their seasonally ephemeral presence. While this eliminates temporal acquisition bias, it removes a recognition channel that is highly discriminative in expert botanical identification. Future editions could incorporate reproductive features through temporally stratified campaigns spanning multiple seasons and geographic sites across Algerian biomes, enabling the study of phenological variation, such as the autumn colour shift in \textit{P.~atlantica} and the deciduous-evergreen contrast among \textit{Quercus} species, as an explicit experimental factor rather than a confounding one.

The evaluation relied on two architectures (DenseNet121 and Swin Transformer) trained with identical hyperparameters and without domain adaptation techniques. Although the strong correlation between architecture rankings (Spearman $\rho \geq 0.94$) indicates that residual variation is primarily data-driven, it remains an open question whether domain adaptation or domain generalisation methods (e.g., adversarial alignment, style transfer, meta-learning) could further reduce the cross-team gap observed under TOTO. Similarly, the CTV framework was applied only to whole-image classification; extending it to object detection, segmentation, or multi-label health assessment would test whether the observed benefits of collaborative training generalise across tasks. These directions, along with validation on external datasets collected independently of the AgrI Challenge, constitute the natural next steps for this line of work.

\FloatBarrier

\section{Conclusion}

This study introduced Cross-Team Validation (CTV) and demonstrated its utility through the AgrI Challenge benchmark, in which 12 teams independently collected 50,673 images of six tree species.

TOTO results established the severity of the single-source generalization problem: mean test accuracies of 81.19\% (DenseNet) and 87.21\% (Swin) compared to validation accuracies of 97.40\% and 98.59\%, with validation-test gaps of 16.20\% and 11.37\% respectively. Cross-team accuracy ranges of 48.2\%--95.3\% (DenseNet) and 65.2\%--98.4\% (Swin) confirm that differences in data collection practices are a primary driver of generalization performance.

LOTO results demonstrated that collaborative multi-team training transforms the generalization landscape. Mean test accuracy increased to 95.31\% ($\pm$3.27\%) for DenseNet121 and 97.04\% ($\pm$2.08\%) for Swin Transformer, improvements of 14.12 and 9.83 percentage points over TOTO. The validation-test gap collapsed from 16.20\% to 2.82\% (DenseNet) and from 11.37\% to 1.78\% (Swin), reductions of 82\% and 84\%. Performance variance was reduced by 40\% (DenseNet) and 54\% (Swin). Under LOTO, 83\% of held-out teams (Swin) and 50\% (DenseNet) exceeded 95\% test accuracy.

The team rankings remained highly consistent across both protocols and both architectures (Spearman $\rho \geq 0.94$), confirming that residual variance under LOTO reflects dataset-level properties. The most disadvantaged team in TOTO (Organization team: 68.32\% DenseNet) gained 25.57 percentage points under LOTO, illustrating how datasets that limit single-source generalization can contribute effectively to a diverse collaborative pool.

Together, these results operationalize the data-centric AI thesis in a realistic field setting: data diversity is the primary determinant of model robustness, and collaborative evaluation via CTV provides a principled means to quantify cross-domain generalization. The AgrI Challenge framework, CTV protocols, and dual-architecture baselines are made available to support future research in robust agricultural AI development.

\section*{Acknowledgements}

The authors would like to sincerely thank all individuals and organizations who contributed to the success of the \textbf{AgrI Challenge 2024}.

We first express our deepest appreciation to \textbf{ATM Mobilis}, our main sponsor, for their generous support and commitment to promoting innovation and youth initiatives. We also gratefully acknowledge \textbf{BK Fire (BKFIRE)} for their valuable sponsorship and contribution to the organization of the event.

We sincerely thank the \textbf{administrations of both ENSIA and ENSA} for their continuous guidance, logistical assistance, and institutional support, which were essential to the realization of this event.

Special thanks go to the organizing clubs from both institutions, \textbf{ENSIA Tech Community (ETC)}, \textbf{GDSC ENSIA}, \textbf{MECA TECH}, and the \textbf{Eco Club}, as well as to all organizers whose dedication ensured the smooth execution of the challenge.

We are also deeply grateful to the \textbf{coaches and mentors} for their expertise, guidance, and support throughout the competition.

Finally, we thank all participants and volunteers whose enthusiasm and engagement made the \textbf{AgrI Challenge 2024} a successful and impactful experience.

\section*{Funding}

The organization of the \textbf{AgrI Challenge 2024} was primarily funded by \textbf{ATM Mobilis}, \textbf{École Nationale Supérieure d’Intelligence Artificielle (ENSIA)}, and \textbf{BK Fire (BKFIRE)}. Their financial contributions were instrumental in supporting the logistical, technical, and operational aspects of the event.

\section*{Data Availability}

The AgrI Challenge dataset is available upon request through the official challenge website. The website provides all relevant information, including the dataset description, structure, data collection methodology, benchmark splits, and evaluation protocols.

\begin{itemize}
\item AgrI Challenge website: \url{https://www.ensia.edu.dz/agri-challenge.html}
\item Source code and training pipeline: \url{https://github.com/Agri-Challenge/agri-challenge-scripts-2024}
\end{itemize}

The repository includes the preprocessing pipeline, duplicate detection scripts, and training configurations necessary to reproduce the experiments reported in this study.

\section*{Declaration of Competing Interest}

The authors declare no competing interests.

\bibliography{references}

\end{document}